\documentclass{article}

\usepackage{microtype}
\usepackage{graphicx}
\usepackage{subfigure}
\usepackage{booktabs} % for professional tables
\usepackage{multirow}
\usepackage{subcaption}

\usepackage{hyperref}
\usepackage{bm}

\usepackage[accepted]{icml2025}

\usepackage{amsmath}
\usepackage{amssymb}
\usepackage{mathtools}
\usepackage{amsthm}
\usepackage{xurl}

\usepackage[capitalize,noabbrev]{cleveref}

\theoremstyle{plain}

\theoremstyle{definition}

\theoremstyle{remark}

\usepackage[textsize=tiny]{todonotes}

\newcommand{\ours}{\textit{SAE-V}}
\newcommand{\old}{\textit{SAE}}
\newcommand{\model}{MLLM}

\icmltitlerunning{\ours{}: Interpreting Multimodal Models for Enhanced Alignment}

\begin{document}

\twocolumn[
\icmltitle{\ours{}: Interpreting Multimodal Models for Enhanced Alignment}



\icmlsetsymbol{equal}{*}

\begin{icmlauthorlist}
\icmlauthor{Hantao Lou}{equal,yyy,comp}
\icmlauthor{Changye Li}{equal,yyy,comp}
\icmlauthor{Jiaming Ji}{yyy,comp}
\icmlauthor{Yaodong Yang}{yyy,comp}
\end{icmlauthorlist}

\icmlaffiliation{yyy}{Institute for AI, Peking University, Beijing, China}
\icmlaffiliation{comp}{State Key Laboratory of General Artificial Intelligence, Institute for AI, Peking University, Beijing, China}

\icmlcorrespondingauthor{Hantao Lou}{hantaolou.htlou@gmail.com}
\icmlcorrespondingauthor{Yaodong Yang}{yaodong.yang@pku.edu.cn}

\icmlkeywords{Machine Learning, ICML}

\vskip 0.3in
]



\printAffiliationsAndNotice{\icmlEqualContribution} % otherwise use the standard text.

\renewcommand{\thefootnote}{*}

\begin{abstract}

% As multimodal large language models (\model{}s) continue to advance, their interpretability and alignment remain significant challenges due to the complexity of integrating different modalities and the inconsistent quality of training data. Existing interpretability methods, such as \old{}, have shown promise but fail to generalize effectively to \model{}s or struggle to provide comprehensive interpretation. Similarly, existing data filtering approaches often require training additional models and are designed for text-only tasks, making them difficult to scale and generalize to multimodal scenarios. In this work, we present \ours{}, a mechanistic interpretability framework that extends the \old{} paradigm to multimodal contexts, enabling the interpretation of multimodal data and models. Moreover, \ours{} facilitates the investigation of multimodal alignment dynamics and improves \model{} alignment through a novel data filtering method that is self-guided and model-intrinsic. Specifically, when applied to the alignment process of LLaVA-NeXT-7B and Chameleon-7B, \ours{}-based data filtering methods could achieve 115\% and 118\% performance with only 50\% and 70\% data, respectively. Our results highlight \ours{}’s ability to improve alignment efficiency while offering interpretable insights into multimodal integration.

With the integration of image modality, the semantic space of multimodal large language models (\model{}s) is more complex than text-only models, making their interpretability more challenging and their alignment less stable, particularly susceptible to low-quality data, which can lead to inconsistencies between modalities, hallucinations, and biased outputs. As a result, developing interpretability methods for \model{}s is crucial for improving alignment quality and efficiency. In text-only LLMs, Sparse Autoencoders (\old{}s) have gained attention for their ability to interpret latent representations. However, extending \old{}s to multimodal settings presents new challenges due to modality fusion and the difficulty of isolating cross-modal representations. To address these challenges, we introduce \ours{}, a mechanistic interpretability framework that extends the \old{} paradigm to \model{}s. By identifying and analyzing interpretable features along with their corresponding data, \ours{} enables fine-grained interpretation of both model behavior and data quality, facilitating a deeper understanding of cross-modal interactions and alignment dynamics. Moreover, by utilizing cross-modal feature weighting, \ours{} provides an intrinsic data filtering mechanism to enhance model alignment without requiring additional models. Specifically, when applied to the alignment process of \model{}s, \ours{}-based data filtering methods could achieve more than 110\% performance with less than 50\% data. Our results highlight \ours{}’s ability to enhance interpretability and alignment in \model{}s, providing insights into their internal mechanisms. \footnote{Our codebase and model are released at \href{https://github.com/PKU-Alignment/SAE-V}{Github} and \href{https://huggingface.co/PKU-Alignment/SAE-V}{Huggingface}.}

% As multimodal large language models (\model{}s) continue to advance, their interpretability and alignment remain significant challenges due to the complexity of integrating different modalities. Current interpretability approaches, including \old{} and neuron-level analysis, have demonstrated
% potential in this area. However, these approaches either fail to migrate to \model{}s directly, or struggle to explain comprehensive \model{}s. In this work, we propose \ours{}, a mechanistic interpretability-based framework for interpreting \model{}s and enhancing their alignment. \ours{} extends the sparse autoencoder paradigm to multimodal contexts, enabling the analysis of joint semantic spaces across modalities. Based on \ours{}, we further interpret multimodal alignment processes and improve \model{} alignment through a novel data filtering method. Especially, when applied to the alignment process of LLaVA-NeXT-7B and Chameleon, \ours{}-based data filtering methods could achieve 115\% and 118\% performance with only 50\% and 70\% data, respectively.
\end{abstract}

\renewcommand{\thefootnote}{\arabic{footnote}}%

\begin{figure}[htbp]
    \centering
    \includegraphics[width=0.45\textwidth]{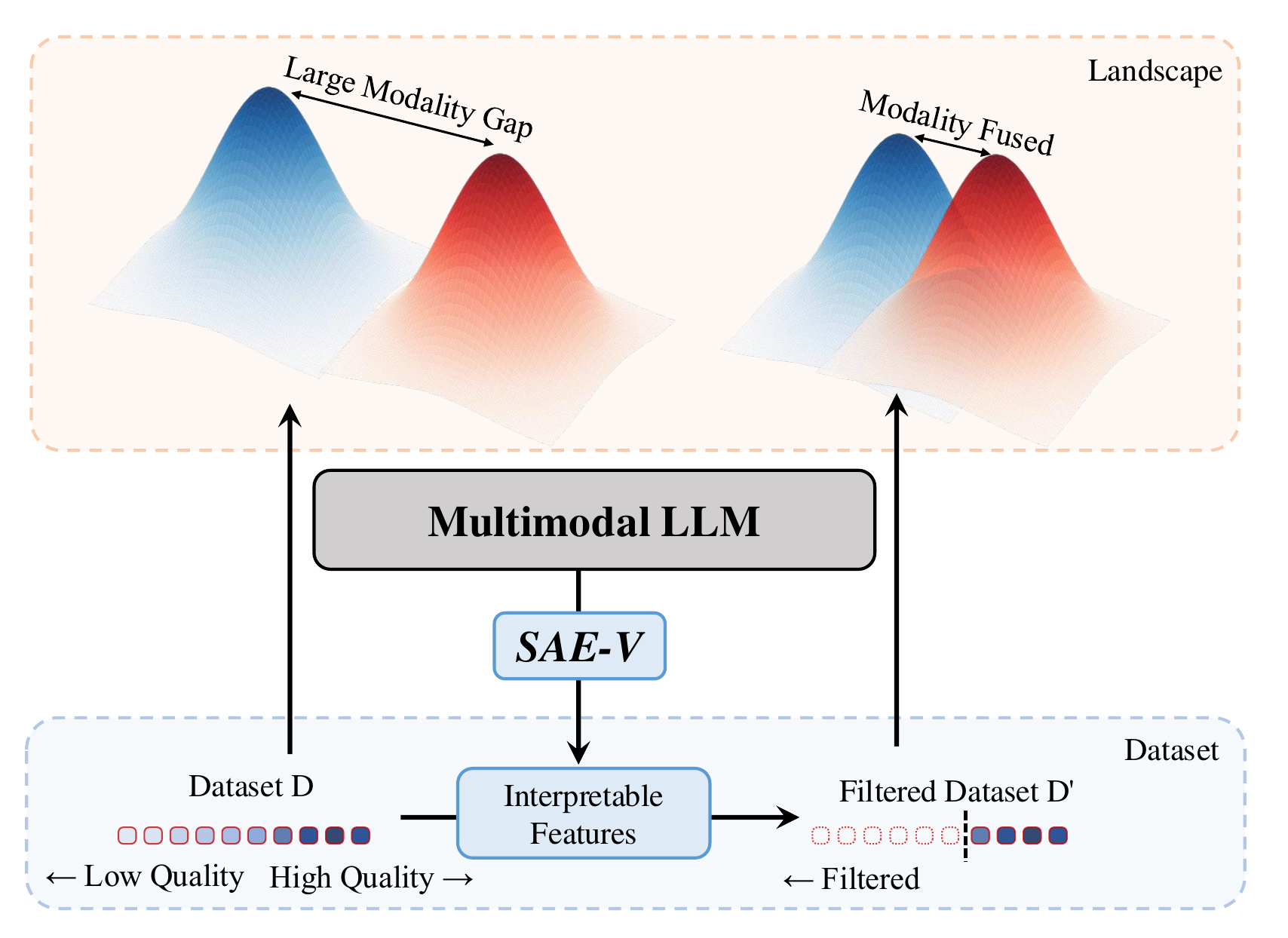}
    \vspace{-0.4cm}
    \caption{\textbf{Operational Dynamics of \ours{} Based Data Filtering Method}. \ours{} encodes and interprets the representation inside \model{} during alignment and inference time. Based on this representation, we could reveal the modality gap within the data, and improve the alignment process through the selection of modality-fused, high-quality data. This pipeline performs data filtering without requiring additional models, relying instead on \model{} itself to prioritize high-value data effectively.}
    \label{fig:main}
\end{figure}

\section{Introduction}
\label{sec:intro}
With the development and success of large language models (LLMs) \citep{dubey2024llama, achiam2023gpt}, researchers have begun to introduce visual understanding to these models, thereby extending their operational scope from language to a mix of vision and language, resulting in the creation of powerful multimodal large language models (\model{}s) \citep{alayrac2022flamingo, liu2024visual, team2024gemini, team2024chameleon}. To enhance the multimodal understanding capabilities of \model{}s, the research community has explored various architectures, including using individual image/text encoders to encode cross-modal information into a joint representation space \citep{zhang2023video, liu2024visual, zhu2024minigpt, wu2024nextgpt} and leveraging image tokenizers to transform all inputs into a unified token sequence \citep{team2024chameleon, xie2024show, wu2024janus, wang2024emu3}. Despite the difference in the architectures of these models, their essential goal is the same: Fuse the text and image representation space into a joint multimodal semantic space.

% 引入两个认知：现有图文模型的图文理解能力并不完美；学界为提升理解能力做了很多尝试，但是并没有揭示背后的原理
As \model{}s continue to scale up in both size and capability, their interpretability and controllability remain a significant challenge \citep{zhang2018visual, stan2024lvlm}. Currently, mechanistic interpretability techniques such as circuit analysis \citep{olsson2022context} and dictionary learning with sparse autoencoders \citep{cunningham2023sparse} are the most widely recognized approaches to interpreting LLMs. However, their application to \model{}s, especially in the context of cross-modal integration, has been limited. There is a pressing need for specialized tools and frameworks that can unravel the intricate workings of \model{}s.

% 另一方面，当前的可解释性尝试很少被用在真实的对齐场景中，这同样使得这些方法的评估变得较为困难。自顶向下的方法，如RepE，activation steering等可以使用control或者unlearning的方式直接对这些可解释性方法对应的控制效果进行评估，而对于自底向上的方法，如circuit，sparse autoencoder，crosscoder等，除了loss之外有效的评估方式较少。因此，我们是否可以提出一种自底向上的多模态可解释性方法，能够直接作用于对齐过程，提升对齐效果？
Moreover, current interpretability efforts are focused mainly on interpreting models, rather than applying this interpretability to real alignment situations, which also makes it difficult to evaluate these methods effectively. Top-down approaches, such as Representation Engineering \cite{zou2023representation} and activation steering \cite{turner2023steering,panickssery2023steering}, can directly evaluate the control effects of interpretability methods through control or unlearning techniques. However, for bottom-up methods like circuit analysis \cite{wang2022interpretabilitywildcircuitindirect}, sparse autoencoders (\old{}s) \cite{cunningham2023sparse}, and cross-coders, effective evaluation methods beyond loss reduction are limited. Based on the previous discussion, \textit{can we propose a bottom-up multimodal interpretability approach that can directly enhance the alignment process?}

%在这份工作中，我们根据文本模态大模型的机制可解释性实践，在视觉-语言模型上开发了一系列可解释性工具，并将这些可解释性工具运用于解释文本模态大模型向视觉-语言模型的训练过程，以及视觉-语言模型的图文能力增强过程。此外，我们利用上述过程中发现的图文联合表征空间的可解释性特征和多模态模型能力的关系，设计了一个基于可解释性工具的数据筛选工具，能够过滤掉数据集中不利于模型发展多模态理解能力的数据，用更少量数据实现更强的对齐效果。总而言之，我们的工作有以下几个contributions:

In this work, we developed \ours{}, a mechanistic interpretability framework for \model{}s that extends the \old{} paradigm to \model{}s. These tools are then applied to interpret the training process of transitioning from LLMs to \model{}s, as well as the process of enhancing the multimodal capabilities of \model{}s. Furthermore, utilizing the interpretable features of \ours{} models and their relationship to \model{} capabilities, we designed a data filtering metric based on \ours{}. This metric can filter out data that hinder the development of multimodal understanding, achieving stronger alignment with a smaller dataset. Overall, our work makes the following contributions:

\begin{figure}[htbp]
    \centering
    \includegraphics[width=\linewidth]{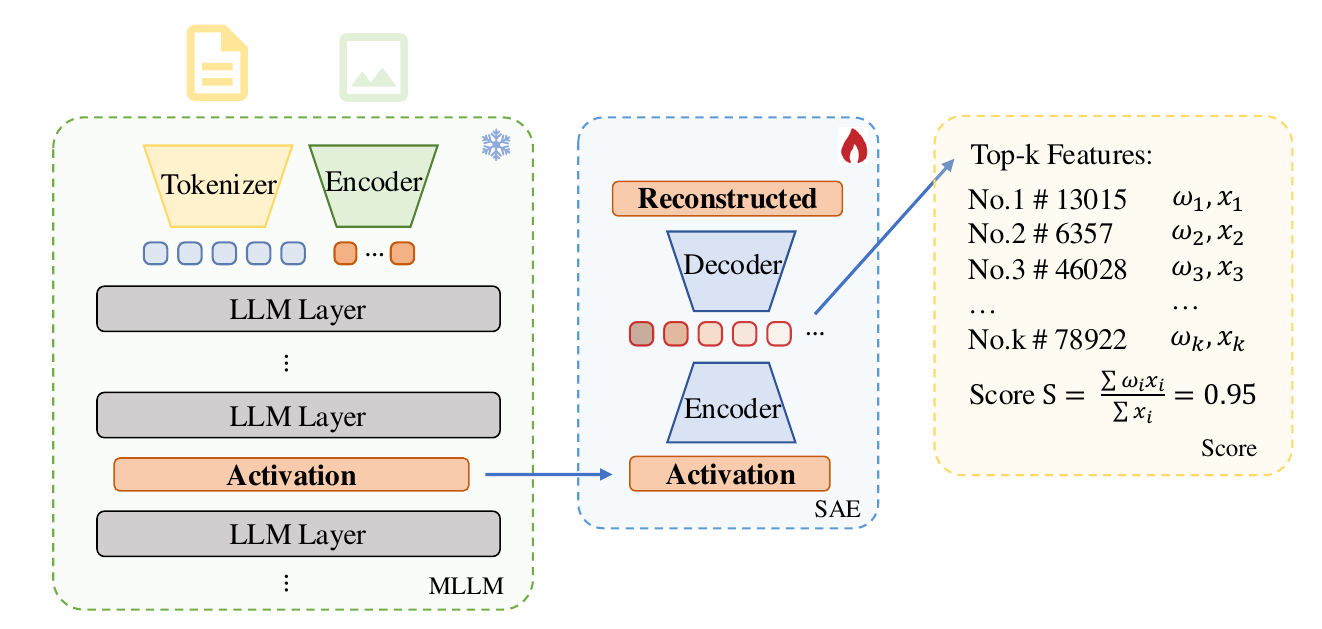}
    \vspace{-0.6cm}
    \caption{\textbf{The interpretability and data filtering pipeline of \ours{}.} \ours{} is trained to encode \model{} activations into sparse, interpretable features. We first acquire the cross-modal weight of these features via \ours{} models, then reversely score the given data by the weighted average of each feature's score. In this way, we provide an intrinsic data filtering tool by eliciting \model{}'s latent representation of these data. }
    \label{fig:pipeline}
\end{figure}

\begin{itemize}
    \item \textbf{Multimodal interpretability tool} We developed mechanistic interpretability tools for \model{}s based on previous attempts on LLMs and trained corresponding \ours{} models. We demonstrated that \ours{} models trained on \model{}s can effectively extract interpretable features, and \ours{} models can be transferred to the corresponding LLMs. Specifically, the reconstruction loss of our \ours{} models trained on \model{}s is 38.3\% and 50.6\% compared to the \old{} model when applied to \model{}s and LLMs, respectively.
    \item \textbf{Interpreting Multimodal Alignment Process} We utilized \ours{} to study the feature distribution throughout the alignment process. We discovered that the feature distribution of \ours{} corresponds to the \model{}’s performance on multimodal understanding tasks.
    \item \textbf{Filtering metric to improve alignment} Based on the previous investigation with \ours{}, we developed a metric to filter multimodal datasets and acquire high-quality data, therefore improving alignment quality and efficiency. Experiments demonstrate that our filtering tool achieves more than 110\% performance compared to the full dataset while using 50\% less data, underscoring the efficiency and effectiveness of \ours{}.
\end{itemize}

\section{Methodology}

In this section, we present our method to train, evaluate, and apply \ours{} to interpret \model{}s and multimodal data.

\subsection{Preliminary: Sparse Autoencoder Paradigm}

We adopt \ours{} (denoted as $\mathcal{S}_\theta$) architecture from the methodology proposed in \cite{bricken2023towards}, which comprises an encoder and a feature dictionary $\mathcal{F}_\theta \text{:}\ \{\bm{f}_k\}_{k=1}^{\bm{n}} $ as a decoder.
Let the input be denoted as $H \in \mathbb{R}^{\bm{l}\times \bm{m}}$, where $\bm{l}$ donates the number of input tokens and $\bm{m}$ donates the shape of hidden state token, the hidden state of a specific layer of a \model\ $\mathcal{M}_\theta$. The \ours{} encoding operation $\mathcal{S}_\theta(\cdot)$ is defined as
\begin{align}
\label{eq:acti}
    Z=\text{ReLU}(H\times W_{\text{enc}}+b_{enc}), 
\end{align}
where $Z \in \mathbb{R}^{\bm{l}\times \bm{n}}$ is the feature activation of the input. The reconstruction loss of $\mathcal{S}_\theta$ donates as
\begin{align}
    \label{eg:reconst}
    \mathcal{L}_R={||H-Z \times (\bm{f}_1, \bm{f}_2, \dots, \bm{f}_n) ^\top||}_2^2.
\end{align}

The training loss is defined by 
\begin{align}
\label{eq:reconst}
    \mathcal{L}=\mathcal{L}_R+\lambda \mathcal{L}_1,
\end{align}
where $\mathcal{L}_1=||Z||_1$ adds a sparsity constraint to the learned features and $\lambda$ is a hyperparameter controlling the level of sparsity. The training results could also quantized by incorporating an additional sparsity constraint via 
$\mathcal{L}_0 = ||Z||_0$, which counts the number of nonzero elements in the learned features $Z$.

\begin{algorithm}[t]
\caption{Cosine similarity score Ranking}
\label{alg:cosim}
\begin{algorithmic}
\STATE {\bfseries Require:} Text token vocabulary: $\mathcal{T}$; vision token vocabulary: $\mathcal{V}$; multimodal dataset $\mathcal{D}=\{\bm{d}_i\}_{i=1}^{p} $; \model\ 
parameterized by $\mathcal{M_\theta}$; \ours{} model $\mathcal{S_\theta}$; features of \ours{} model $\mathcal{F_\theta}\text{:}\ \{\bm{f}_k\}_{k=1}^{n}$; activation bound\text{:} $\delta$; cosine similarity function \textit{C} (Equation~\ref{eg:cosi})
\STATE {\bfseries Ensure:} Ranked data $\mathcal{D}_R$
\STATE \textcolor{blue}{\textbf{Stage 1: Collect Feature Activation Token}} \
\STATE Initialize activated token set of features $\mathcal{A}_k \gets \varnothing $ \;
\STATE $\mathcal{D}_s \gets \textit{Sample}(\mathcal{D})$ \;
\FOR {each $\bm{d}_i \in \mathcal{D}_s$}
    \STATE $H_i \gets \mathcal{M_\theta}(\bm{d}_i)$
    \STATE $Z_i \gets \mathcal{S_\theta}(H_i)$ \;
    \FOR {each $\bm{f}_k \in \mathcal{F_\theta}$}
        \STATE $\mathcal{A}_k \gets \mathcal{A}_k \cup \{ \bm{h}_j \text{:}\ \bm{h}_j \in H_i,\bm{z}_j = \bm{e}_j Z_i, \bm{z}_{jk} > \delta\}$
    \ENDFOR
\ENDFOR
\STATE \textcolor{blue}{\textbf{Stage 2: Compute Cross-modal Weight}} \;
\STATE Initialize cross-modal weight of features $\mathcal{\omega}_k \gets 0 $ \;
\FOR {each $\bm{f}_k \in \mathcal{F_\theta} $} 
% \IF{$\mathcal{A}_k \neq \varnothing$}  
    \STATE $\mathcal{\omega}_k \gets \textit{C}(\textit{TopK}(\mathcal{A}_k \cap \mathcal{T}),\textit{TopK}(\mathcal{A}_k \cap \mathcal{V}))$ 
% \ENDIF
\ENDFOR
\STATE \textcolor{blue}{\textbf{Stage 3: Rank Dataset by Cross-modal Weight}} \;
\STATE Initialize cross-modal score of data $\bm{s}_i \gets 0 $ \;
\FOR {each $\bm{d}_i \in \mathcal{D}$}
     \STATE $Z_i \gets \mathcal{S_\theta}(\mathcal{M_\theta}(\bm{d}_i))$ \;
    \STATE $F_i \gets \{\bm{f}_k \text{: } \exists\ z_j \in Z_i ,\ \mathrm{ s.t.}\ z_j\  \text{activates } \bm{f}_k\}$
    \STATE $\bm{s}_i \gets \sum_{\bm{f}_k \in F_i} \omega_k$
\ENDFOR 
\STATE $\mathcal{D}_R \gets \textit{Sort}(\mathcal{D}, \{\bm{s}_i\}_{i=1}^n)\ $
\end{algorithmic}
\end{algorithm}

\subsection{Interpreting Multimodal Data with \ours{}}
\label{sec:datafilter}

It has been previously demonstrated~\cite{gao2024scaling,cunningham2023sparse} that \old{} can be employed to interpret how LLMs encode semantic information from these models. This feature motivates us to apply \ours{} to assess the quality of the data and thus facilitate data filter for alignment.

We adopt a cosine similarity score ranking algorithm for data filtering (shown in Algorithm~\ref{alg:cosim}). Let the multimodal training dataset be donated as $\mathcal{D}=\{\bm{d}_i\}_{i=1}^{p} $, where $\bm{d}_i=(u_{\bm{1}},u_{\bm{2}},\dots,u_{\bm{l}})$, $ u_{\bm{j}} \in \mathcal{T} \lor u_{\bm{j}} \in \mathcal{V}, \bm{j}=\bm{1},\bm{2},\dots,\bm{l}$, is a sequence of tokens from text vocabulary $\mathcal{T}$ and tokens from vision vocabulary $\mathcal{V}$. We acquire feature activation token $\bm{z}_j$ by \model\ forward and Equation~\ref{eq:acti}, \textit{i.e.}
\begin{align}
\label{eg:feat}
    H_i&=\mathcal{M}_\theta (\bm{d_i}) \\
    Z_i&=\mathcal{S}_\theta (H_i), \\
\label{eg:l0}
    \bm{z}_j&=\bm{e}_j Z_i , \\
    \text{where}\ \bm{e}_j &= (0, 0, \ldots, \underbrace{1}_{j\text{-th position}}, \ldots, 0). \nonumber
\end{align}
We define a \ours{} feature $\bm{f}_k$ is activated on $\bm{z}_j \ \textit{if}\ \bm{z}_{jk}>\delta$, where $\delta$ is activation bound. Correspondingly, we state that $\bm{f}_k$ is activated on $\bm{d}_i\ \textit{if}\  \exists\ \bm{z}_j \in Z_i,\ \mathrm{s.t.}\ z_j\ \text{activates}\ \bm{f}_k $.

Our algorithm~\ref{alg:cosim} consists of three stages: (1) \emph{Collecting feature activation tokens from dataset}, (2) \emph{Computing cross-modal weight of \ours{} features}, and (3) \emph{Ranking dataset by cross-modal weight}.

\begin{enumerate}
    \item \textbf{Feature Activation Token Collecting}  
    We first sample a small subset $\mathcal{D}_S$ of the training dataset $\mathcal{D}$ and input these samples into the \model\ to obtain hidden states $H$. These hidden states are then fed into the \ours\ encoder to extract feature activations defined as Equation~\ref{eg:feat}. For each feature, we collect its hidden state tokens thereby obtaining a sample of feature activation tokens across the dataset.
    
    \item \textbf{\ours{} Feature Weighting}  
    For each feature $\bm{f}_k$, we identify its top-K hidden state text tokens $\bm{t} = \textit{TopK}(\mathcal{A}_k \cap \mathcal{T})$ and top-K hidden state vision tokens $\bm{v} = \textit{TopK}(\mathcal{A}_k \cap \mathcal{V})$, where the top-K is ranked by the activation value $\bm{z}_{jk}$ of the token. We then compute the cosine similarity between the two lists of tokens, donating the cross-modal weight of feature $\bm{f}_k$ as
    \begin{align}
    \label{eg:cosi}
        Cosine(\bm{t},\bm{v})=\mathbb{E}_{i,j} \frac{\bm{t}_i \cdot \bm{v}_j}{||\bm{t}_i||||\bm{v}_j||},
    \end{align}
    % $\text{where}\  \bm{t}_i \in \textit{TopK}(\mathcal{A}_k \cap \mathcal{T})\ \text{and}\ \bm{v}_i \in \textit{TopK}(\mathcal{A}_k \cap \mathcal{V})$.
    % We classified these features according to the tokens they activated in. 
    which represents the capability of the feature to capture multimodal information within data.
    
    \item \textbf{Data Ranking}
    Using the weighted features of \ours{} model, we score the entire training dataset. The cosine similarity score of each piece of data is defined as the sum of the cosine similarity scores of its activating features. We rank the data set by the score and the resulting cosine similarity score order allows us to filter data that are better aligned with the structures of multimodal semantic information.
\end{enumerate}

We present our experiments and results in Section~\ref{sec:Alignment}, demonstrating that our cosine similarity score ranking method can effectively filter high-quality data from the training data set.

\section{Interpretability Analysis with \ours{}}
% \section{\ours{} provides a key to Multimodality}

\begin{figure*}[htbp]
    \centering
    \begin{minipage}[b]{0.48\textwidth}
        \centering
        \includegraphics[width=\textwidth]{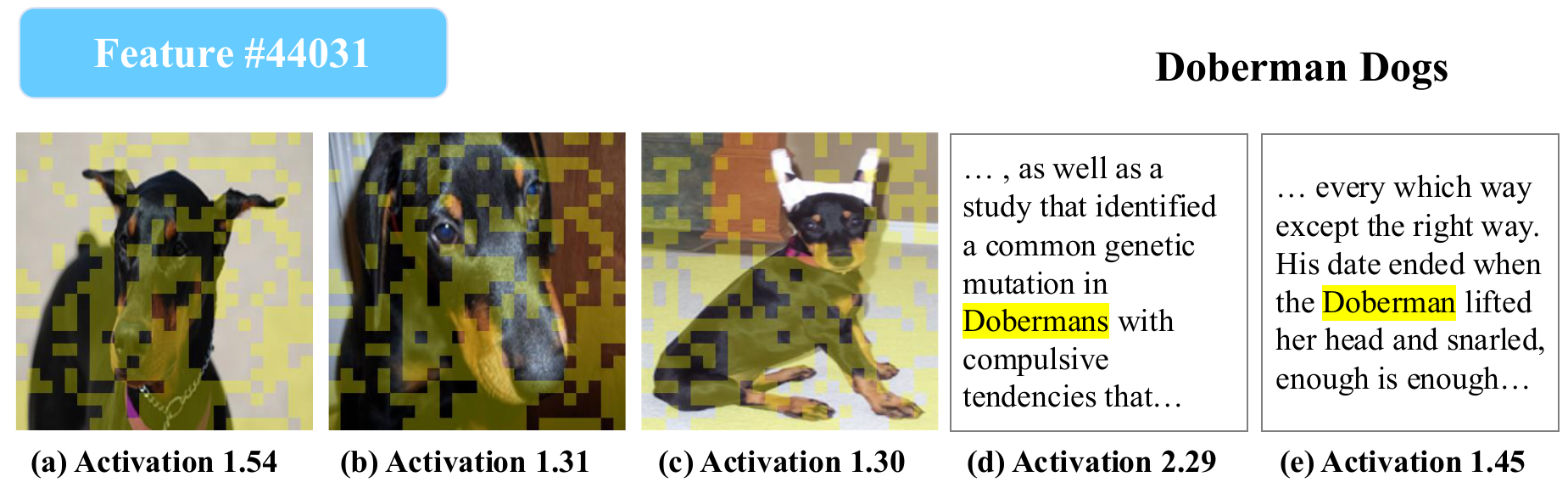}
        \\
        \small (a)
    \end{minipage}
    \hfill
    \begin{minipage}[b]{0.48\textwidth}
        \centering
        \includegraphics[width=\textwidth]{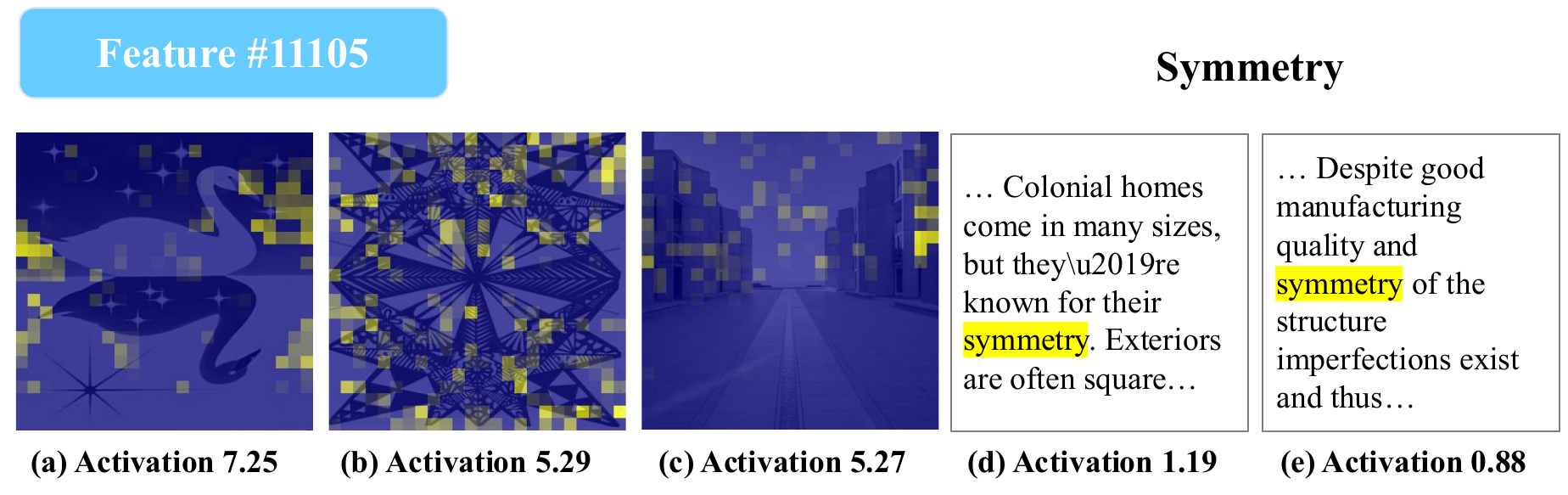}
        \\
        \small (b) 
    \end{minipage}
    \vspace{-0.2cm}
    \caption{\textbf{Examples of interpretable features discovered by \ours{}.} We presented examples of interpretable \ours{} features on LLaVA-NeXT-7B that demonstrate cross-modal semantic consistency. \textbf{(a)} Feature \#44031 exhibits strong activation for `Doberman dogs' across both text and image modalities, showing \ours{}'s ability to identify specific concepts with concrete physical meanings. \textbf{(b)} Feature \#11105 captures the abstract concept of `Symmetry' across different modalities, activating for various symmetry patterns including left-right, top-bottom, and central symmetry, with activation regions precisely aligned with symmetrical elements in images. These examples illustrate that \ours{} can discover features representing both concrete entities and abstract concepts that maintain consistent semantic meaning across modalities, going beyond what traditional probing methods on raw activations can achieve.}
    \label{fig:interp_example}
\end{figure*}

In this section, we conduct experiments on the \ours{} paradigm, aiming to demonstrate the capability and transferability of \ours{} model. We also performed experiment to prove the effectiveness of our \ours{}-based data interpreting tool from the inference side.

\subsection{Training and Evaluating \ours{} Model}

We trained a series of \old{} and \ours{} models on \model{}s and their base LLMs. We evaluated the performance of these models, and the results demonstrated that \ours{} model is capable of interpreting \model{}s and that \ours{} model trained on \model{} can be effectively transferred to its original LLM.

\subsubsection{Experiment Setup}
\label{sec:setup}

\paragraph{Datasets} For text-only and multimodal situations, we selected the Pile \cite{gao2020pile} and Obelics \cite{laurencon2023obelics} datasets separately. Specifically, we sampled 100K data from each dataset as the train set and 10K data as the test set.
%我们在这两个数据集中各抽样了100k数据作为训练集，10k数据作为测试集
The Pile is a diverse language modeling dataset for LLM pretraining, and Obelics is a massive interleaved image-text dataset for \model{} pretraining. These two datasets are widely recognized in various pretraining and interpretability works \cite{black2022gpt, biderman2023pythia, cunningham2023sparse, team2024chameleon}. 

\paragraph{Models} We selected two generic \model{}s, LLaVA-NeXT models (including LLaVA-NeXT-Mistral-7B, LLaVA-NeXT-Vicuna-7B and LLaVA-NeXT-Vicuna-13B) \cite{liu2024visual} and Chameleon-7B \cite{team2024chameleon}, as our target models. These models represent two distinct architectures, and testing our method on them can demonstrate that our method is applicable to different architectures.
% \begin{itemize}
%     \item LLaVA-NeXT-7B is a powerful \model{} using CLIP vision tower to convert images into representations, it's fine-tuned with multimodal data based on Mistral-7B. 

%     \item Chameleon-7B is a foundational mixed-modal early-fusion model using a separate image tokenizer to transform images into discrete tokens. It's pretrained with a mix of text-only data and multimodal data to enable image understanding and generation simultaneously.
% \end{itemize}

Additionally, we also studied Anole-7B \cite{chern2024anole} and Mistral-7B \cite{jiang2023mistral} to compare the behavior of \old{} and \ours{} models before and after fine-tuning, specifically the transitioning fine-tuning from LLM to \model{}. Anole-7B is a variant of Chameleon-7B, with its image generation capability unlocked, while Mistral-7B is the base LLM of LLaVA-NeXT-7B. \footnote{We present the detailed training setup and hyper-parameters in \textit{Appendix~\ref{appendix:training_parameters}}.}

\paragraph{Evaluation Metrics} To evaluate the performance of \ours{} models, we use two key metrics: $\mathcal{L}_0=||\bm{z}||_0$ where $\bm{z}$ is defined in Equation~\ref{eg:l0} and reconstruction loss $\mathcal{L}_R$ in Equation~\ref{eg:reconst}. $\mathcal{L}_0$  quantifies the number of activated features, reflecting the method's ability to extract interpretable features, while reconstruction loss measures the method's activation reconstruction capability compared with the model output, indicating the method's accuracy in giving interpretations.

\subsubsection{Experiment Result}

\begin{table}[ht]
\centering
\small
\begin{tabular}{lll}
\toprule
\textbf{Model} & \textbf{Method}  & \textbf{$\mathcal{L}_0$ }  \\ 
\midrule
\multirow{2}{*}{\shortstack{\textbf{Chameleon-7B}}} 
& \old{} & 24757.6 \\ 
& \ours{} &  50.1  \\ 
\midrule
\multirow{2}{*}{\shortstack{\textbf{Anole-7B}}} 
& \old{} & 62.1  \\ 
& \ours{} & 50.1  \\ 
\midrule
\multirow{2}{*}{\shortstack{\textbf{LLaVA-NeXT-Mistral-7B}}} 
& \old{} &   94.5 \\ 
& \ours{} &   192.5  \\ 
\midrule
\multirow{2}{*}{\shortstack{\textbf{LLaVA-NeXT-Vicuna-7B}}} 
& \old{} &   3162.96 \\ 
& \ours{} &   585.64  \\ 
\midrule
\multirow{2}{*}{\shortstack{\textbf{LLaVA-NeXT-Vicuna-13B}}} 
& \old{} &   128.56 \\ 
& \ours{} &   193.63  \\

\bottomrule
\end{tabular}
\caption{\textbf{The $\mathcal{L}_0$ metric of \old{} and \ours{} models}. $\mathcal{L}_0$ indicates the sparsity cost (average activated feature number). The results vary significantly across models due to their architectural differences. For Anole-7B and Chameleon-7B, \ours{} models maintain lower $\mathcal{L}_0$, suggesting more efficient feature utilization. However, LLaVA-NeXT models shows a contrary pattern with \ours{} requiring higher feature activation than \old{}. We propose that extra activated features of \ours{} model are introduced by extra vision tokens in multimodal data. Notably, Chameleon-7B and LLaVA-NeXT-Vicuna-7B with \old{} model exhibits an unusually high sparsity cost, attributed to multiple unseen vision tokens in the inference stage.}
\label{tab:sae2saev}
\end{table}

\begin{figure}[ht]
    \centering
    \includegraphics[width=\linewidth]{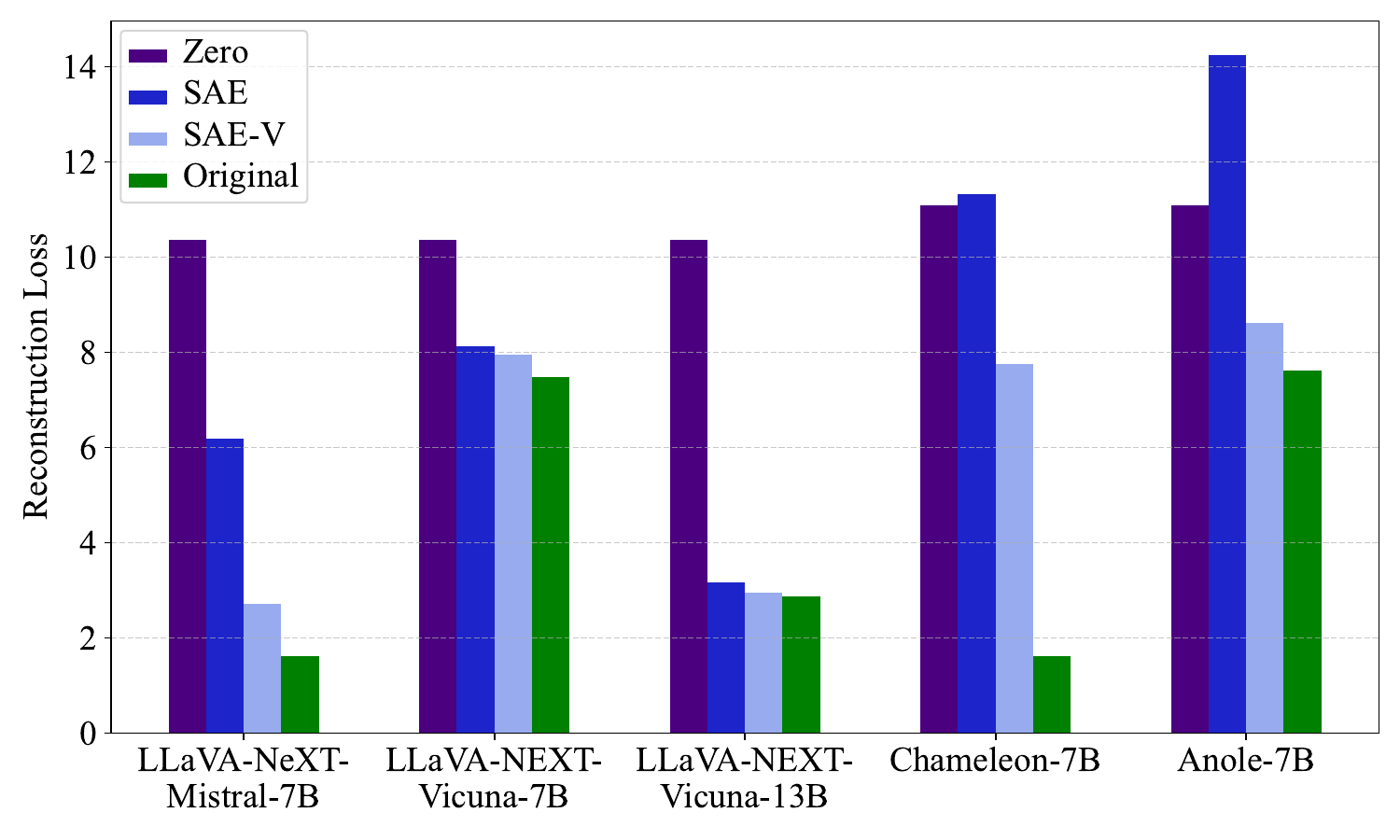}
    \vspace{-0.5cm}
    \caption{\textbf{Reconstruction capability of \old{} and \ours{} models.}  Each section compares the metrics of zero (set all activations as zero), \old{} model, \ours{} model, and the Original reference state. \ours{} model consistently demonstrates superior reconstruction performance across all tested models. In Chameleon-7B and Anole-7B, \old{} performs worse than the zero baseline, which indicates that \old{} trained in text data fails to capture interpretable features in these \model{}s.}
    \label{fig:l0}
\end{figure}

\paragraph{Capability of \ours{} Model} We compare the performance of \ours{}  and \old{} on different multimodal models. The $\mathcal{L}_0$ (shown in Table \ref{tab:sae2saev}) varies significantly across the three models. For LLaVA-NeXT-7B, the $\mathcal{L}_0$ of \ours{} is much higher than that of \old. For Chameleon-7B, \ours{} performs normally, whereas the $\mathcal{L}_0$ of \old{} is abnormally high, indicating that \old{} fails to extract sparse features. We suppose that the failure is attributed to a large number of unseen vision tokens for \old{} during the inference stage. For Anole-7B, the $\mathcal{L}_0$ of \old{} and \ours{} are nearly identical. The reconstruction loss (shown in Figure~\ref{fig:l0}) of \ours{} is lower than \old{} and is closer to the original activation, demonstrating that \ours{} behaves much better at reconstructing original activation than \old{} across all three models. The results indicate that \ours{} outperforms \old{} in terms of capability.

\begin{figure}[ht]
    \centering
    \includegraphics[width=\linewidth]{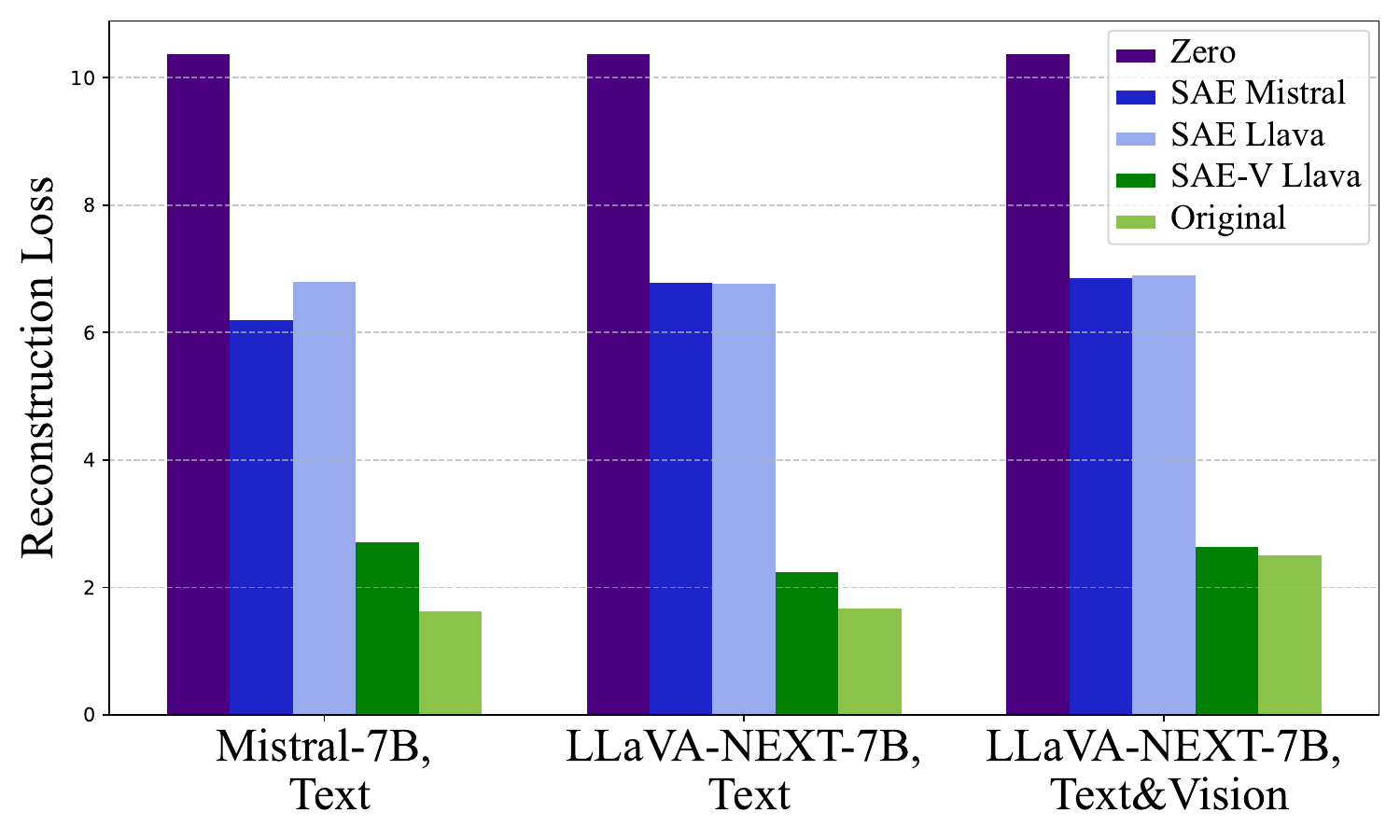}
    \vspace{-0.5cm}
    \caption{\textbf{Reconstruction performance of \old{} and \ours{}.} The x-axis shows different models and task configurations, text indicates text-only task, and text \& vision indicates multimodal task. The colored bars represent five experiment groups (zero activation, \old{} of Mistral-7B, \old{} of LLaVA-NeXT-7B trained with the Pile, \old{} of LLaVA-NeXT-7B trained with Obliecs, original performance). Across various settings, \ours{} consistently demonstrates superior transferability compared to \old{} and achieves reconstruction loss close to the original performance, the maximum relative gap being 67.28\%. \old{} based on Mistral-7B and \old{} based on LLaVA-NeXT-7B achieves nearly the same loss in all tasks and models, indicating the equivalence of training \old{} with \model{} and its base LLM.}
    \label{fig:sae_convert}
\end{figure}

\paragraph{Transferability of \ours{} Model} The transferability of \old{}s between foundation models and instruction-tuned models has been extensively investigated in text-only contexts \citep{kissane2024saes,kutsyk2024do,gallifant2025sparse}, as it demonstrates whether \old{}s can capture universal semantic features within LLMs. Similarly, the transferability from \model{}s to corresponding LLMs serves as a critical metric for the quality of features learned by \ours{}. We compared the reconstruction performance of \ours{} model trained on LLaVA-NeXT-7B and \old{} model to prove that \ours{} model trained on \model{}s can generalize to its base LLM. The findings (shown in Figure~\ref{fig:sae_convert}) indicate that across different settings, \ours{} model consistently achieves the best performance. Moreover, when trained on both \model{} and LLM, \old{} model exhibits nearly identical reconstruction loss values, showing its robust transferability.

These results highlight that training \ours{} model for \model{}s with multimodal data is effective for interpreting \model{}s, and even LLMs, as \old{} model trained solely on textual data fail to extract and disentangle the hidden representations of \model{}s effectively. Moreover, \ours{} model demonstrates superior capability in reconstructing the reasoning features of \model{}s compared to the standard \old{} model. 
%In terms of transferability, \ours{} achieves the best performance across both \model{}s and their original models.

\subsection{Apply \ours{} Model on Multimodal Data}
% \subsection{\ours{} captures pivotal insights in Multimodal Data}
\label{sec:Semantic}

\begin{figure}[h!]
    \centering
    \includegraphics[width=0.8\linewidth]{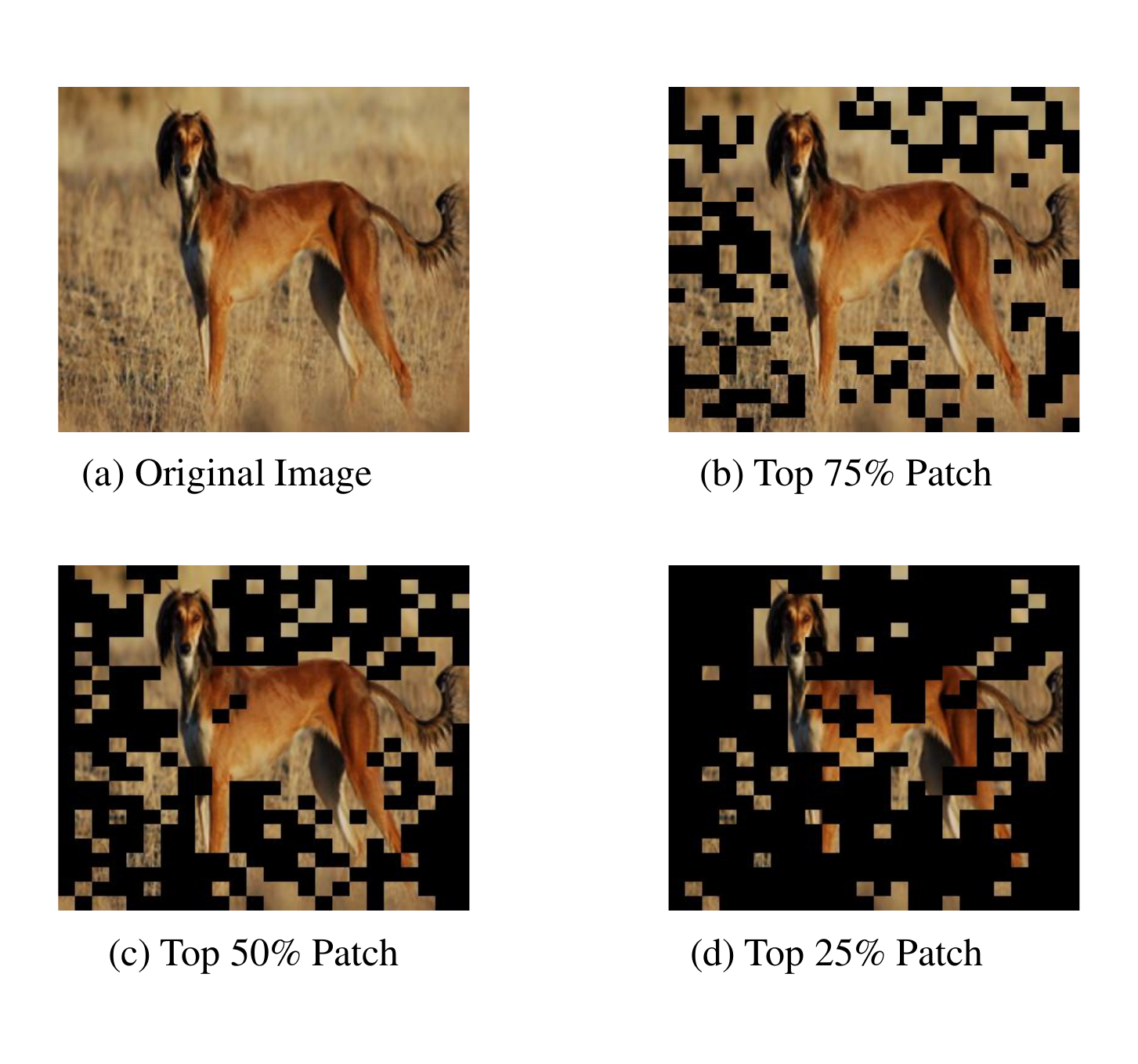}
    \vspace{-0.6cm}
    \caption{\textbf{Case analysis of image patch filtering using $\mathcal{L}_0$ metric.} We rank and filter image patches according to the number of features activated on them. The top row shows the original image (a) and its reduced-patch versions retaining 75\% (b), 50\% (c), and 25\% (d). In this dog image, the patches are filterd out from edge to the middle and preserved almost only dog patches, suggesting that \ours{} model is preserving the main semantic information of the image.}
    \label{fig:seminar_experi_1}
\end{figure}

In this section, we conduct an image classification task on the ImageNet dataset \cite{imagenet15russakovsky} to investigate whether \ours{} can capture the key information within images and to validate the effectiveness of the methods proposed in Section~\ref{sec:datafilter} on multimodal data. We apply 4 methods, namely $\mathcal{L}_0$, $\mathcal{L}_1$, co-occurring $\mathcal{L}_0$, and cosine similarity score, where co-occurring $\mathcal{L}_0$ is defined as the number of features activated on at least one text and image token. The cosine similarity score is defined as the sum of cross-modal weights of features, consistent with Algorithm~\ref{alg:cosim}.
We adopt these metrics to filter image patches, thus obtaining images that preserve 75\%, 50\%, and 25\% patches, respectively. \footnote{We present complete algorithms of 4 methods in Appendix~\ref{appendix:algo}.}

\begin{figure}[h!]
    \centering
    \includegraphics[width=0.4\textwidth]{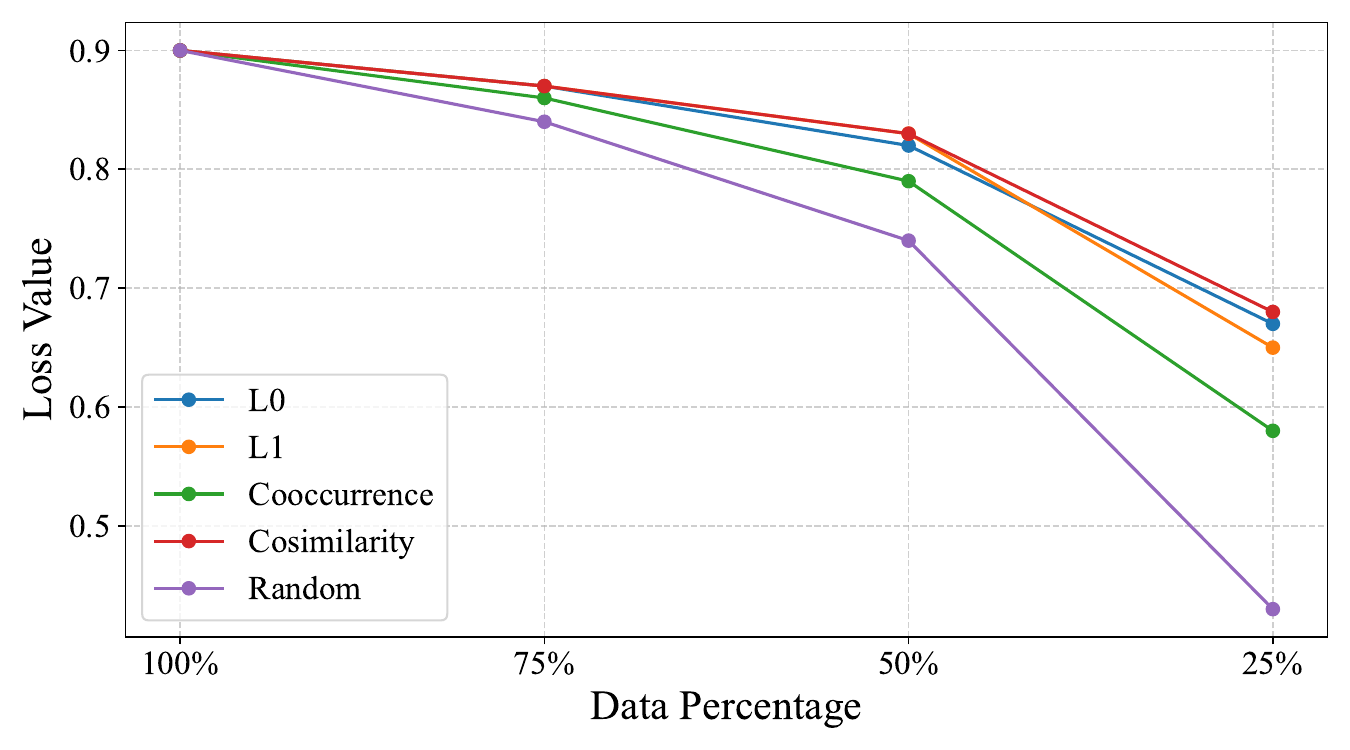}
    \vspace{-0.4cm}
    \caption{\textbf{The classification performance on ImageNet.} We compare the classification accuracy after filtering the image patches with $\mathcal{L}_0$ method, $\mathcal{L}_1$ method, co-occurrence $\mathcal{L}_0$ method, cosine similarity score method, and the random baseline. All methods achieve high accuracy when preserving 75\% or 50\% patches and $\mathcal{L}_0$ method, $\mathcal{L}_1$ method, and cosine similarity score method maintains high accuracy even in the least patches. The result shows that \ours{} is efficient in capturing critical information from images.}
    \label{fig:seminar_experi_2}
\end{figure}

\paragraph{Case Analysis} Figure~\ref{fig:seminar_experi_1} illustrates the image patches when using $\mathcal{L}_0$\ metric for filtering. Even when employing the simplest $\mathcal{L}_0$ metric, \ours{} is still able to effectively capture the critical semantic information of the image. \footnote{More cases and analyses are presented in Appendix~\ref{appendix:Semantic_Experiment}.}

\paragraph{Quantized Results} The quantized results are presented in Figure~\ref{fig:seminar_experi_2}, where we observe that, for all methods, preserving 75\% or 50\% of the patches achieves an accuracy close to that obtained using the full image. In the challenging scenario where only 25\% of the patches are retained, $\mathcal{L}_0$ method, $\mathcal{L}_1$ method, and cosine similarity score method maintain accuracy levels close to 70\%, and all methods significantly surpass the accuracy obtained by the random preservation method. These results demonstrate that \ours{} can accurately capture critical information in images and that the methods proposed in Section \ref{sec:datafilter} effectively utilize \ours{} features during inference. \footnote{More experiments and their quantative analysis are presented in Appendix ~\ref{appendix:quant_anal}}

\section{Alignment Experiment}
\label{sec:Alignment}

In this section, we adopt cosine similarity score rank-
ing algorithm as a data filter (as shown in Algorithm \ref{alg:cosim}) to acquire high-quality data for model alignment.

\subsection{Experiment Setup}

\begin{figure*}[htbp]
    \centering
    \begin{minipage}[b]{0.32\textwidth}
        \centering
        \includegraphics[width=\textwidth]{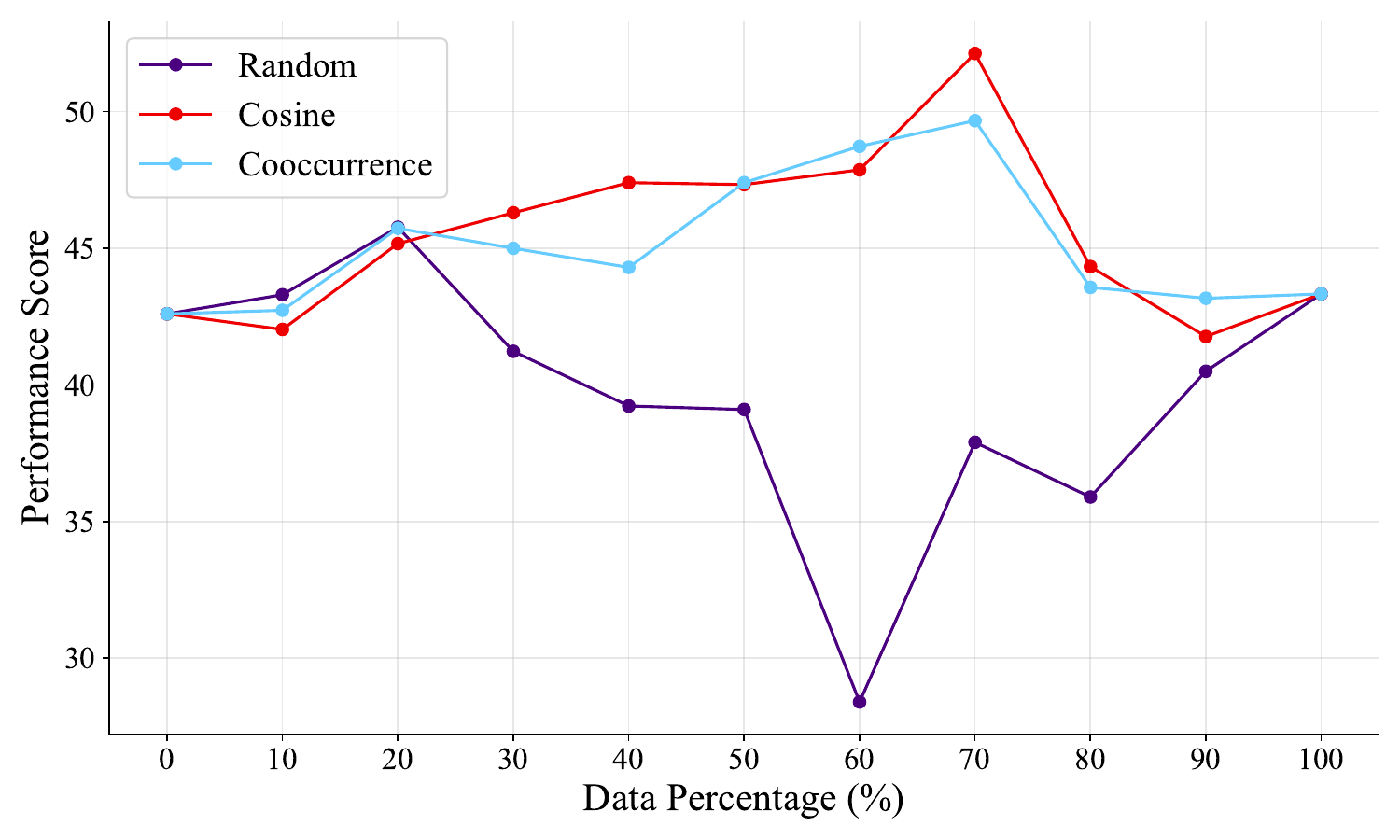}
        \\
        \small (a) Performance on Align-Anything dataset
        \label{fig:chameleon_aa}
    \end{minipage}
    \hfill
    \begin{minipage}[b]{0.32\textwidth}
        \centering
        \includegraphics[width=\textwidth]{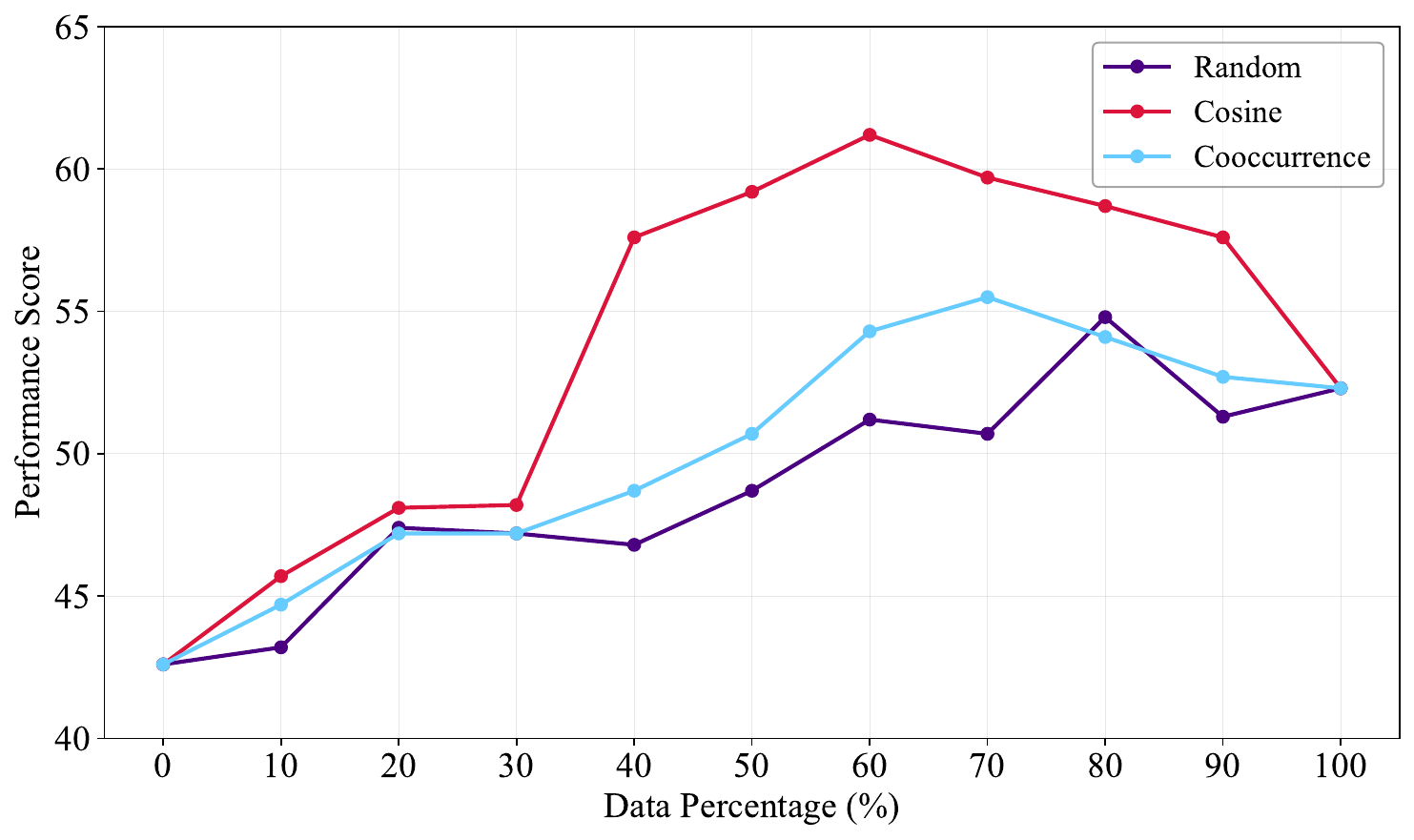}
        \\
        \small (b) Performance on MMInstruct dataset
        \label{fig:chameleon_mminstruct}
    \end{minipage}
    \hfill
    \begin{minipage}[b]{0.32\textwidth}
        \centering
        \includegraphics[width=\textwidth]{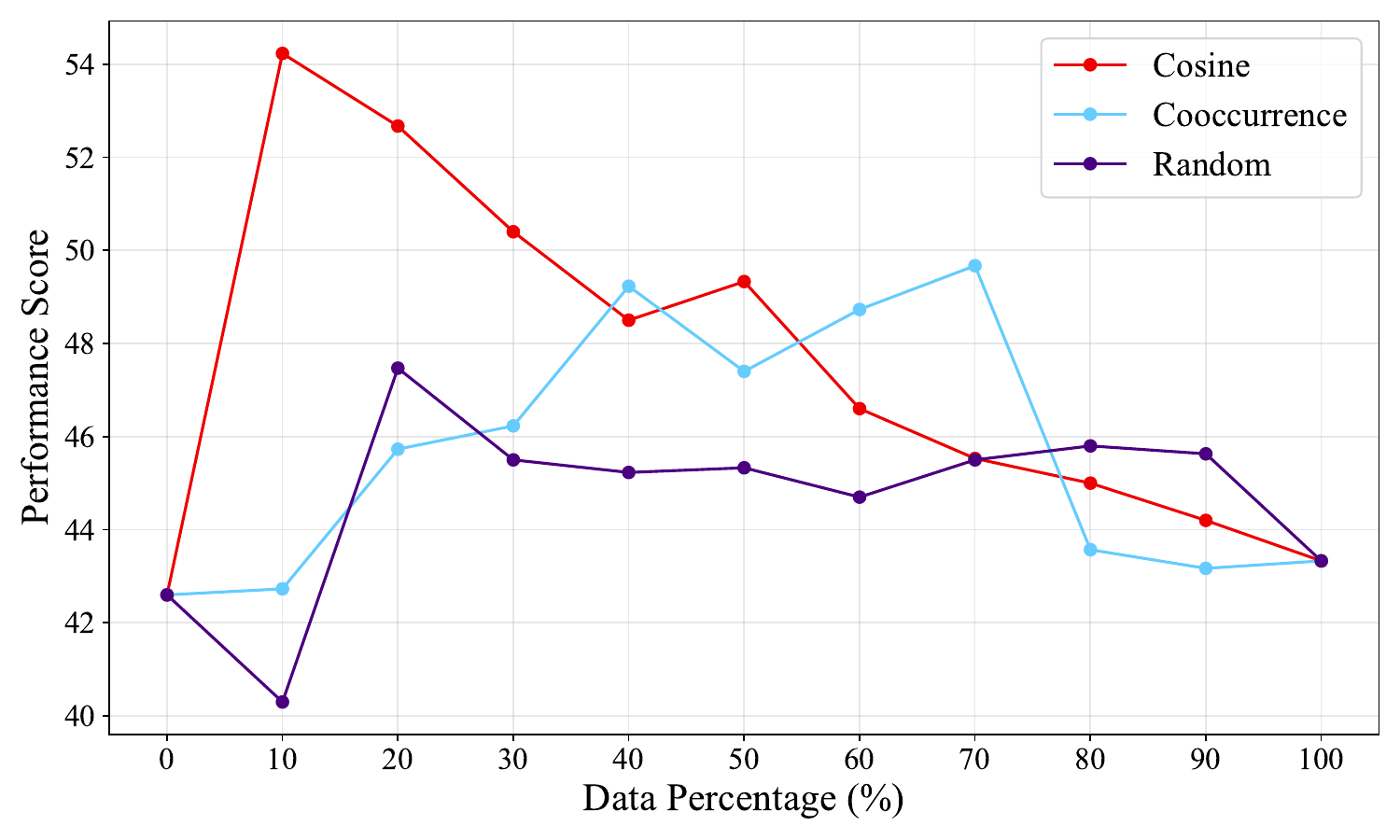}
        \\
        \small (c) Performance on RLAIF-V dataset
        \label{fig:chameleon_rlaifv}
    \end{minipage}
    \vspace{-0.2cm}
    \caption{\textbf{Ablation study of \ours{}-based data filter method on Chameleon-7B model and different datasets.} We replicate \ours{}-based data filter on Chameleon-7B model and three distinct datasets. The results show that \ours{}-based data filter method is still effective on different datasets and different model architectures and image encoding methods other than CLIP. \textbf{(a)} On Align-Anything dataset, both cosine similarity and co-occurrence filters perform better than random filters at almost every data percentage, and the cosine similarity filter achieved a score of 52.1 (120\% of the full dataset's performance) with 70\% data. \textbf{(b)} On MMInstruct, the cosine similarity filter outperforms the random filter baseline at every data proportions. Specifically, the cosine similarity filter achieved a score of  61.2 (117\% of the full dataset's performance) with 60\% data. \textbf{(c)} On RLAIF-V dataset, the cosine similarity filter achieves the highest score of 54.2 (125\% of the full dataset's performance) with only 10\% data, demonstrating its supreme efficiency.}
    \label{fig:chameleon_ablation}
\end{figure*}
\paragraph{Dataset and Model} Consistent with Section~\ref{sec:setup}, we selected LLaVA-NeXT-7B \cite{liu2024visual} and Chameleon-7B \cite{team2024chameleon} for our alignment experiment. Since the LLaVA-NeXT-7B model is rather powerful in multimodal capabilities, we selected the Align-Anything \cite{ji2024align} text-image-to-text dataset for our experiment. Align-Anything is a 400K multimodal preference dataset containing fine-grained annotated multimodal input-output preference data, and we used the 40K subset of text-image input and text output in our experiment.

\paragraph{Algorithm} We adopt the cosine similarity score ranking algorithm (shown in Algorithm \ref{alg:cosim}) as a filter to exclude data with low scores. In addition, we also adopt two algorithms, the $\mathcal{L}_0$ ranking, and the co-occurrence ranking. \footnote{Detailed descriptions of these ablation algorithms and their corresponding hyperparameters are provided in Appendix~\ref{appendix:align}.}

\paragraph{Evaluation} To evaluate the efficiency of our methods, we applied Direct Preference Optimization (DPO) to the model using the filtered datasets. \footnote{We present detailed training parameters in Appendix~\ref{appendix:modeltrain}.} We then evaluate the multimodal capabilities of the model using LLaVA-Bench \cite{liu2024visual} benchmarks. 

\begin{figure}[h]
    \centering
    \includegraphics[width=0.9\columnwidth]{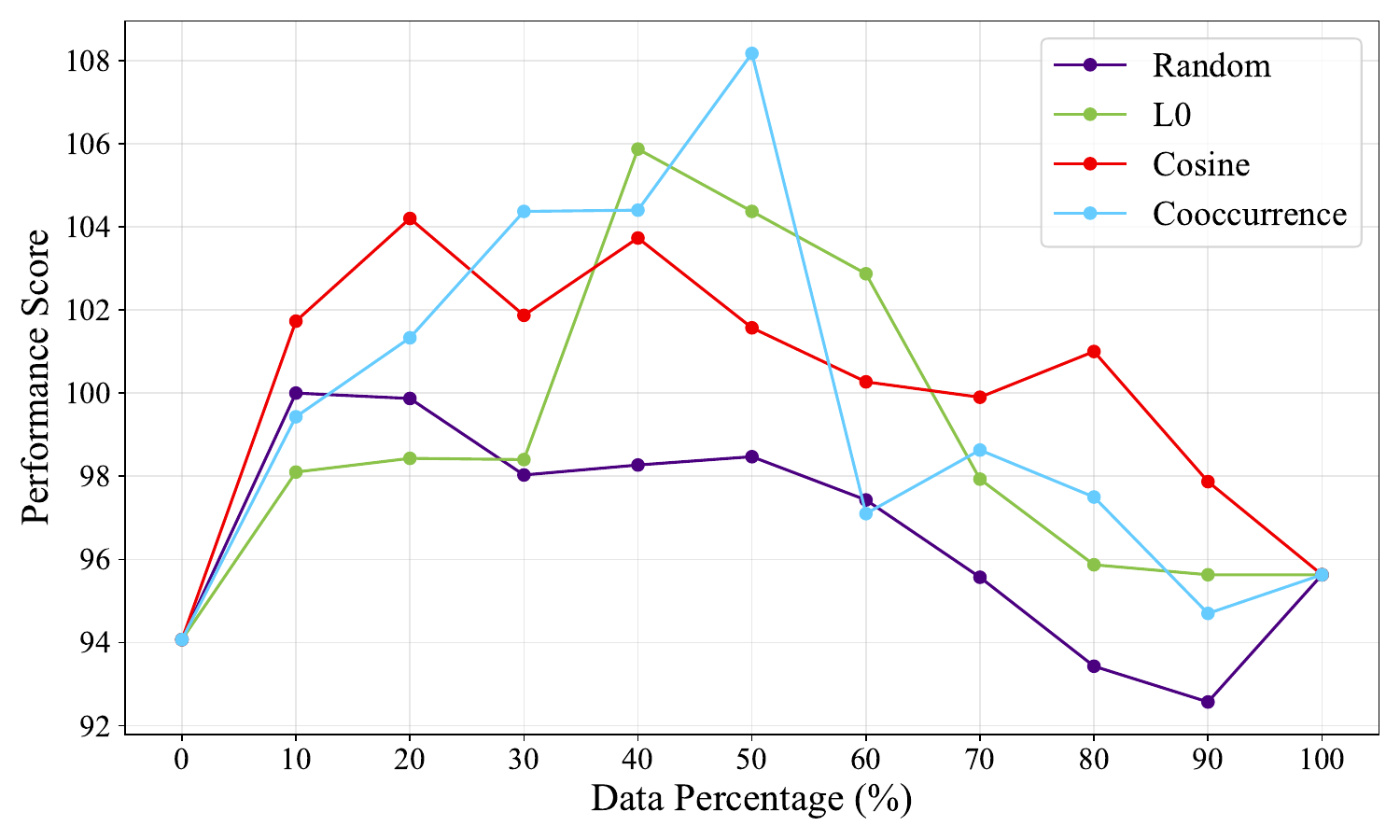}
    \vspace{-0.4cm}
    \caption{\textbf{The performance of \ours{}-based data filter method} We evaluated \ours{}-based data filter method on LLaVA-NeXT-7B model, Align-Anything dataset, and LLaVA-Bench benchmark. The results show that all SAE-V-based methods significantly outperform the random selection baseline, while the cosine similarity filter achieved 108\% of the full dataset’s performance with only 20\% of the data, and the co-occurrence filter peaked at 50\% of the data, reaching a score of 108.17. As a more straightforward utilization of \ours{} model, the L0 filter also generally outperforms the random selection baseline.}
    \label{fig:alignment}
\end{figure}
\subsection{Experiment Results}
\label{sec:alire}

%写结果或者discussion时可以考虑使用asset/static_plot_gradient和asset/static_plot_original这两张图

We performed \ours{}-based data filter with different filtering ratios on the LLaVA-NeXT-7B model and Align-Anything dataset. The filtered datasets were then used to fine-tune \model{}s, which were evaluated on LLaVA-Bench. The results (shown in Figure \ref{fig:alignment}) demonstrate that our \ours{}-based filtering method effectively enhances the alignment of LLaVA-NeXT-7B, even with reduced data. Since most of the data in Align-Anything contribute positively to model alignment, the performance of the model is higher than the base model without any fine-tuning in most cases. At any data filter proportion, the \ours{}-based data filtering method outperforms the random selection baseline, with the best result being 108.17 (115\% of the full dataset’s performance) achieved using 50\% filtered data from the cooccurrence filter, and 104.20 (108\% of the full dataset’s performance) achieved using 20\% filtered data from the cosine similarity filter. However, as the dataset inevitably contains some low-quality data, the performance is optimal with a moderate data proportion and shows a downward trend as the data proportion increases.

\subsection{Relationship between \model{} Capability and \ours{} Features}
\label{sec:cross-modal}

\begin{figure}[ht]
    \centering
    \includegraphics[width=0.9\linewidth]{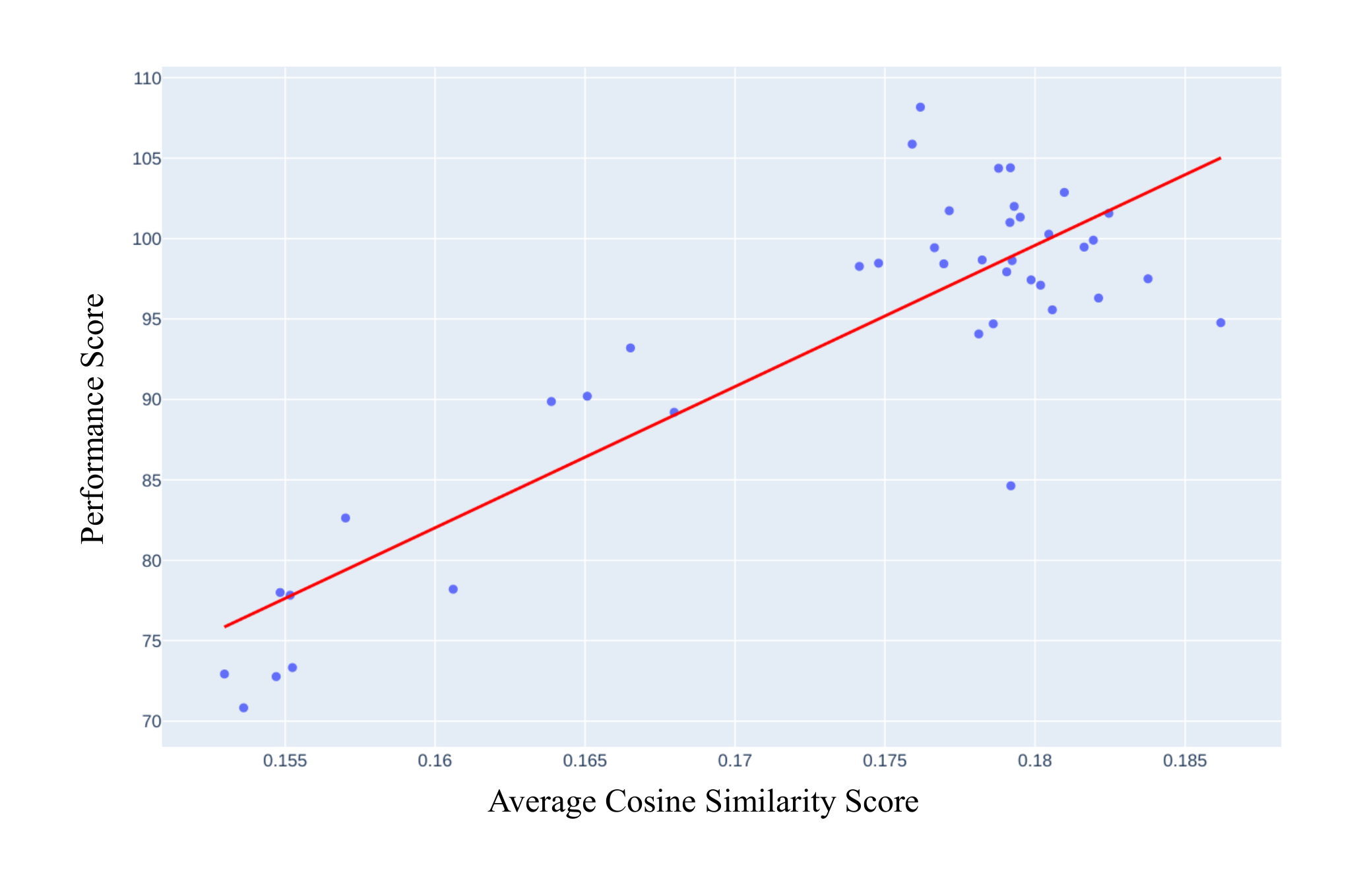}
    \vspace{-0.6cm}
    \caption{\textbf{The relationship between average cosine similarity score and \model{} performance.} We measure the average cosine similarity score of models in Section \ref{sec:alire}, and fit a linear relationship between model performance and average cosine similarity score. The correlation coefficient of the correlation reaches 0.84, suggesting that higher similarity scores on \ours{} features correspond to enhanced \model{} performance.}
    \label{fig:interpcross-modality}
\end{figure}

In the previous section, we demonstrated the effectiveness of utilizing the cosine similarity score for data filters in model training. To further investigate the relationship between model performance and cross-modal similarity, as measured by cosine similarity of \ours{} features, we further measure the average cosine similarity score of these models. Given a dataset, we apply the cosine similarity score ranking algorithm (shown in Algorithm \ref{alg:cosim}) to the \model{}, and we define the \model{}'s average cosine similarity score as the mean score of all non-zero cross-modal weight \ours{} features.

We calculated the average cosine similarity scores for the models discussed in Section~\ref{sec:alire}. The result (shown in Figure \ref{fig:interpcross-modality}) revealed a positive correlation between the average cosine similarity score of \ours{} feature and the performance of \model{}, suggesting that higher similarity scores of \ours{} features correspond to enhanced \model{} performance.

%在上一部分我们发现了cosi algortithm筛选数据训练模型的卓越效果，为了研究这些模型的cosine similarity这样的跨模态相似性和模型表现的关系。给定数据集，我们将一个\model{}运用到 the cosine similarity score ranking algorithm (as shown in Algorithm \ref{alg:cosim})中，我们定义一个\model{}的average cosine similarity score为所有\ours{} feature (不为零)cross-modal weight的平均值。

% 我们计算了section 4.2中的模型的average cosine similarity score，得到Figure~\ref{fig:interpcross-modality}. 我们拟合观察到A positive correlation between the average cosine similarity score and the performance score, which suggests that higher similarity scores are associated with better model performance。

\begin{table*}[t]
\centering
\begin{tabular}{lccccccccccc}
\hline
\textbf{Method} & \textbf{0\%} & \textbf{10\%} & \textbf{20\%} & \textbf{30\%} & \textbf{40\%} & \textbf{50\%} & \textbf{60\%} & \textbf{70\%} & \textbf{80\%} & \textbf{90\%} & \textbf{100\%} \\
\hline
SAE-V & 94.2 & 98.2 & 106.8 & 114.9 & 114.5 & 114.8 & 112.9 & 112.3 & 111.0 & 109.5 & 98.5 \\
\hline
Random & 94.2 & 96.5 & 97.0 & 98.3 & 95.3 & 93.7 & 96.5 & 98.8 & 98.2 & 96.8 & 98.5 \\
\hline
\end{tabular}
\caption{\textbf{The performance of \ours{}-based data filter method on 1/20 Align-Anything dataset} The performance of \ours{}-based data filter method on 1/20 Align-Anything dataset closely resembles that in Figure \ref{fig:alignment}, confirming that alignment metrics on a small validation set resemble the distribution on the complete dataset, enabling efficient hyperparameter selection.}
\label{tab:micro_exp}
\end{table*}

\begin{table}[htbp]
\centering
\small
\begin{tabular*}{\columnwidth}{@{\extracolsep{\fill}}lcccccc@{}}
\toprule
\multirow{2}{*}{Method} & \multicolumn{6}{c}{Performance} \\
\cmidrule{2-7}
& 0\% & 20\% & 40\% & 60\% & 80\% & 100\% \\
\midrule
SAE-V & 104.6 & 105.8 & \textbf{116.7} & 112.3 & 112.0 & 111.2 \\
Random & 104.6 & 105.3 & 105.8 & 107.2 & 110.4 & 111.2 \\
\bottomrule
\end{tabular*}

\caption{\textbf{The performance of SAE-V-based data filter method
on LLaVA-NeXT-Vicuna-13B} We replicate \ours{}-based data filter on LLaVA-NeXT-Vicuna-13B and Align-Anything dataset. The result shows that \ours{}-based data filter outperforms the random selection baseline, and reached the highest performance of 116.67 with 40\% data. This further indicates that \ours{}-based data filter could work regardless of model architecture and sizes.}
\label{tab:ablation_models}
\end{table}

\subsection{Ablation Study}
\label{subsec:alignment_ablation}

% \begin{figure}[ht]
%     \centering
%     \includegraphics[width=0.9\columnwidth]{assets/ablation_model_fig.pdf}
%     \vspace{-0.4cm}
%     \caption{\textbf{The performance of \ours{}-based data filter method on Chameleon-7B} We replicate \ours{}-based data filter on Chameleon-7B model and Align-Anything dataset. The result shows that on Chameleon-7B model, \ours{}-based data filter also achieved significant performance gain compared to the random filter. Specifically, both cosine similarity and co-occurrence filters perform better than random filters at almost every data percentage, and the cosine similarity filter achieved a score of 52.13 (120\% of the full dataset's performance) with 70\% data. This result prove that \ours{}-based data filter method is still effective on different model architectures and image encoding methods other than CLIP.}
%     \label{fig:ablation_models}
% \end{figure}
\paragraph{Ablation on Models} To prove that \ours{}-based data filter method could generalize to distinct model architectures, we replicate \ours{}-based data filter on Chameleon-7B and LLaVA-NeXT-Vicuna-13B models with the Align-Anything dataset. The result (shown in Figure \ref{fig:chameleon_ablation} (a) and Table \ref{tab:ablation_models}) demonstrates that \ours{}-based filter method still shows its effectiveness regardless of model architecture, sizes, and image encoding methods. 

When using a smaller data proportion, the performance is strongly correlated with the data quantity, and thus, the differences between methods are minimal. However, with a larger data proportion, the \ours{}-based filter method significantly surpasses the random filter, achieving a peak of 52.13 (120\% of the performance on full dataset) with 70\% data on Chameleon-7B and 116.67 (104\% of the performance on full dataset) with 40\% data onLLaVA-NeXT-Vicuna-13B. The largest performance gap is observed in the 40-80\% data range, while the differences converge again as the data proportion approaches 100\%. This proves that \ours{}-based data filter is effective on architectures other than CLIP-based \model{}, and shows its potential to generalize across a wide range of models.

% \begin{figure}[htbp]
%     \centering
%     \includegraphics[width=0.9\columnwidth]{assets/chameleon_fig.pdf}
%     \vspace{-0.4cm}
%     \caption{\textbf{The performance of \ours{}-based data filter method on RLAIF-V dataset.} We performed an ablation study with the RLAIF-V dataset and Chameleon-7B model to prove that \ours{}-based data filter method can generalize across datasets. Both cosine similarity and co-occurrence filters generally outperform the random filter on the RLAIF-V dataset. Specifically, the cosine similarity filter achieves the highest score of 54.23 (125\% of the full dataset's performance) with only 10\% data, demonstrating its supreme efficiency.}
%     \label{fig:ablation_datasets}
% \end{figure}

\paragraph{Ablation on Datasets} We also performed an ablation study on the datasets to be filtered. Since the LLaVA-NeXT-7B model is highly capable, most datasets fail to further enhance its multimodal abilities. Therefore, we selected the RLAIF-V \citep{yu2024rlaifv} and MMInstruct \citep{liu2024mminstruct} datasets with the relatively weaker Chameleon-7B model for the dataset ablation study. The results (shown in Figure \ref{fig:chameleon_ablation} (b) and Figure \ref{fig:chameleon_ablation} (c)) further confirm that \ours{}-based data filter is working across different datasets. Moreover, on RLAIF-V, the cosine similarity filter could achieve a score of 54.23 (125\% of the full dataset's performance) by using only 10\% of the data, and on MMInstruct, the cosine similarity filter could achieve a score of 61.2 (117\% of the full dataset's performance) with 60\% data, demonstrating exceptional efficiency.

\paragraph{Ablation on Dataset Sizes} On real deployment scenarios, due to the large size of datasets, it is often difficult to use the \ours{}-based data filter method to search for filtering ratio parameters on the complete dataset. Therefore, we reproduced the \ours{}-based data filter method on a 1/20 subset of the Align-Anything dataset to demonstrate that in practical deployment situations, it is feasible to search for filter ratios on small-scale dataset subsets for application to the complete dataset. The results (shown in Table \ref{tab:micro_exp}) indicate that the performance of \ours{}-based data filter on the small-scale dataset exhibits extremely similar trends with respect to filtered data ratios as those observed on the complete dataset (shown in Figure \ref{fig:alignment}), which demonstrates that searching for filtered data ratios through small-scale dataset subsets is feasible.

\begin{figure}[ht]
    \centering
    \includegraphics[width=0.9\columnwidth]{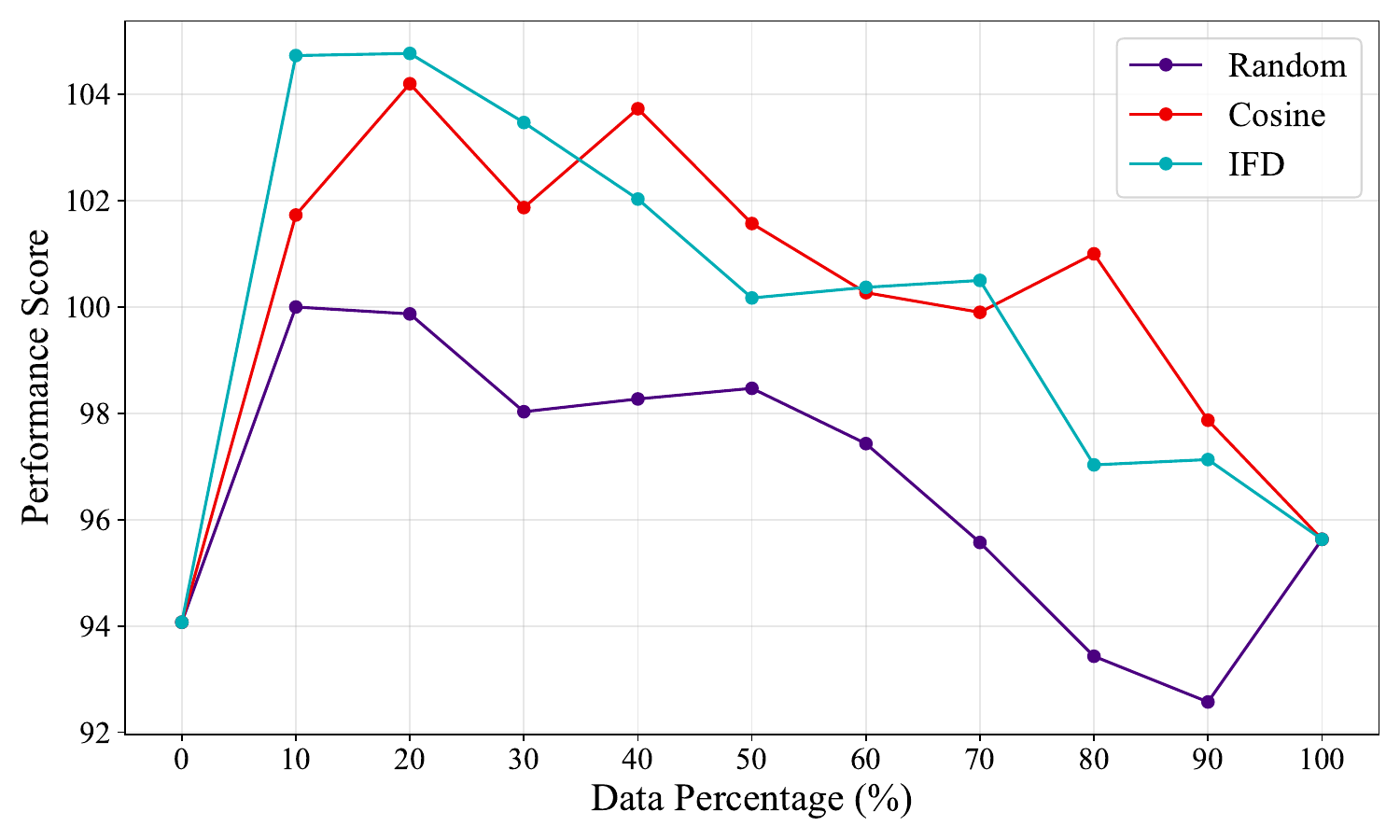}
    \vspace{-0.4cm}
    \caption{\textbf{The performance of \ours{}-based data filter and IFD metrics} We replicate the IFD metric on \model{}s and compare the result with our cosine similarity filter. The results show that although the peak performance of \ours{}-based data filtering is a little lower compared to IFD metric, our method could achieve a generally similar performance compared to IFD metric without introducing additional models and training process.}
    \label{fig:ablation_filter_methods}
\end{figure}

%可以解释一下选择这个数据集的理由是我们没有找到multimodal的filter算法，在单模态中这个算法是最合适的
%写结果或者discussion时可以考虑使用asset/static_plot_gradient和asset/static_plot_original这两张图
\paragraph{Comparation with Other Filtering Methods}
To validate the effectiveness of our \ours{}-based data filtering method, we conducted an ablation study comparing it with other similar data filtering approaches. Since there are currently no widely recognized data filtering methods specifically designed for multimodal data, we adapted the IFD metric \cite{li2023quantity} method to the multimodal setting. The result (shown in Figure \ref{fig:ablation_filter_methods}) suggests that our data filter method achieves a performance comparable to the IFD metric. However, considering that the IFD metric needs to train an additional \textit{cherry model}, our \ours{}-based data filter could directly fit various datasets, demonstrating greater generalizability and efficiency.

\section{Related Work}
\label{sec:Related Work}

\paragraph{Multimodal Large Language Model}

\model{} is a type of LLM integrated with multimodal modules that incorporate multimodal information to deal with multimodal tasks. Based on the method of integrating vision features into the model, most \model{}s can be categorized into three types:
% Some state-of-the-art \model{}s have already achieved significant capabilities in multimodal understanding and generation, such as GPT-4o \cite{OpenAI2024gpt4o}, Gemini-1.5 \cite{team2024gemini}, and Claude3-opus \cite{anthropic2024claude}. 

\begin{itemize}
    \item \textit{CLIP-based \model{}s}: These models encode images with CLIP \cite{radford2021learning} and use MLP to project visual features. Examples include LLaVA \cite{liu2024visual} series and NExT-GPT \cite{wu2023next}.
    \item \textit{Early-Fusion \model{}s}: These models directly tokenize visual features for input. Examples include Chameleon \cite{team2024chameleon} and Janus \cite{wu2024janus} series.
    \item \textit{Q-Former-based MLLMs}: These models use a structure similar to Q-Former \cite{li2023blip} to extract visual representations, represented by Qwen-VL \cite{Qwen-VL} and MiniGPT-4 \cite{zhu2024minigpt}.
\end{itemize}

Our study focuses on the CLIP-based and early-fusion \model{}s. Specifically, we select LLaVA-NeXT-7B and Chameleon-7B as the target models.

\paragraph{Mechanistic interpretability with Sparse Autoencoder}

Mechanistic interpretability seeks to uncover and explain the internal mechanisms that enable models to understand input data and generate responses \cite{rai2024practical}. Specifically, most current mechanistic interpretability methods focus on analyzing features, smaller units that contribute to performing explainable semantic tasks, within models \cite{olah2020zoom}.
% In terms of vision features interpretability, \citet{dravid2023rosetta} discovered the presence of 'Rosetta Neurons' across multiple models, which suggests that different neural networks share similar representations.

Sparse Autoencoder (\old{}) aims to learn sparse and interpretable features from polysemantic model representations \cite{yun2021transformer,bricken2023towards,sharkey2022features,peigne2023features, elhage2022toy}. By introducing sparsity constraints, the activation values in the hidden layers of \old{} are mostly zero, allowing \old{} to encode polysemantic features in LLM to monosemantic ones. 
\citet{zhang2024largemultimodalmodelsinterpret}  firstly attempted to apply \old{} to analyze the open-semantic features of \model{}s.
% \citet{cunningham2023sparse} firstly utilized \old{} to decipher interpretable features from small LLMs. \citet{gao2024scaling} trained an \old{} on GPT-4, proposing a recipe for training \old{}s on LLM and measures to evaluate \old{} metrics. 

In this paper, we extended the scope of \old{} to \model{}s, thereby building \ours{}. We further demonstrated \ours{}'s capability and transferability on \model{}s, and built a data filter tool based on \ours{} to enhance multimodal alignment.

\paragraph{Data Filter in Alignment}

Data filtering ensures that only relevant high-quality data are used during the alignment of LLMs or \model{}s, thus reducing the quantity of data while achieving greater performance \cite{zhou2023lima,chen2023alpagasus,du2023mods,li2023one,li2023quantity,tu2024resofilter}. For example, LIMA \cite{zhou2023lima}, ALPAGASUS \cite{chen2023alpagasus}, and IFD \cite{li2023quantity} use human annotation, API annotation and train a new model for annotation to score data separately. Our method, \ours{}-based data filter, provides a self-guided and interpretable metric to evaluate the similarity of multimodal data, which indicates their qualities. The method is stable and efficient for models of different architectures.
% For example:

% \begin{itemize}
%     \item LIMA \cite{zhou2023lima} emphasizes the importance of quality over quantity in data filtering, demonstrating that effective data filtering can unlock the latent capabilities of pretrained LLMs.
%     \item ALPAGASUS \cite{chen2023alpagasus} integrates ChatGPT into the data filter pipeline, achieving superior performance while significantly reducing training time.
%     \item IFD \cite{li2023quantity} is a self-guided metric to evaluate the alignment difficulty of data samples. The IFD score measures the degree of assistance that instructions provide in generating aligned outputs.
% \end{itemize}

%self guide
%stable
%interp
%efficiency

\section{Conclusion}

This work introduced \ours{}, a framework that extends \old{} to \model{}s and improves their alignment. Through experiments on LLaVA-NeXT-7B and Chameleon-7B, we demonstrated that \ours{} model demonstrates excellent capability and transferability in interpreting \model{} and multimodal data, and \ours{}-based data filtering methods could achieve more than 110\% performance with less than 50\% data. These results highlight \ours{}’s potential to enhance multimodal model interpretability and alignment efficiently.

\paragraph{Limitation} While \ours{} introduces significant advancements in interpreting multimodal models and enhancing alignment through mechanistic analysis, several limitations remain unaddressed and warrant further exploration: (1) Although \ours{} demonstrates superior interpretability and data filtering efficiency compared to \old{}, the theory behind \ours{}, especially the mathematical relationship between image-text similarity metrics, cross-modal co-occurrence features, and model performance, is not fully revealed. (2) Due to resource constraints, \ours{} is primarily evaluated on text and vision modalities, leaving its effectiveness on other modalities such as audio, video, and embodied AI systems unexplored. Our future work will focus on establishing a comprehensive theoretical foundation for \ours{} and extending its application to additional modalities, such as audio, video, and embodied AI systems, to broaden its utility and impact.

\clearpage
\section*{Impact Statement}
The source code and checkpoints of \ours{} mentioned in this paper will be released under the CC BY-NC 4.0 license. This research has several potential risks that must be considered. The interpretability tools introduced in this work, while beneficial for alignment, could also be leveraged to manipulate or reverse-engineer model behaviors in unintended ways. Additionally, while \ours{} provides a self-guided filtering mechanism, it remains dependent on the initial dataset quality, meaning biases in the dataset could still propagate into the final model. We strongly condemn any malicious use of the \ours{} code and checkpoints and advocate for its responsible and ethical use.

\section*{Acknowledgement}

This work is sponsored by National Natural Science Foundation of China (62376013, 623B2003).

\bibliography{main}
\bibliographystyle{icml2025}

\clearpage
\appendix
\section{Details of Interpretability Experiment}

\subsection{Hyperparameter of \old{} and \ours{} Models Training}
\label{appendix:training_parameters}

\begin{table}[ht]
    \centering
    \scriptsize
    \begin{tabular}{lcc}
        \toprule
        \textbf{Hyper-parameters} & \shortstack{\textbf{\old{} and \ours{} of}\\\textbf{LLaVA-NeXT/Mistral}} & \shortstack{\textbf{\old{} and \ours{} of}\\\textbf{Chameleon/Anole}} \\
        \midrule
        \multicolumn{3}{c}{ \textbf{Training Parameters}} \\
        \hline
        total training steps & 30000 & 30000 \\
        batch size & 4096 & 4096 \\
        LR & 5e-5 & 5e-5 \\
        LR warmup steps & 1500 & 1500 \\
        LR decay steps & 6000 & 6000 \\
        adam beta1 & 0.9 & 0.9 \\
        adam beta2 & 0.999 & 0.999 \\
        LR scheduler name & constant & constant \\
        LR coefficient & 5 & 5 \\
        % lp norm & 1.0 & 1.0 \\
        seed & 42 & 42 \\
        dtype & float32 & float32 \\
        buffer batches num & 32 & 64 \\
        store batch size prompts & 4 & 16 \\
        % use ghost grads & False & False \\
        feature sampling window & 1000 & 1000 \\
        dead feature window & 1000 & 1000 \\
        dead feature threshold & 1e-4 & 1e-4 \\
        \hline
        \multicolumn{3}{c}{\textbf{\old{} and \ours{} Parameters}} \\
        \hline
        hook layer & 16 & 8 \\
        input dimension & 4096 & 4096 \\
        expansion factor & 16 & 32 \\
        feature number & 65536 & 131072 \\
        % b-dec init method & zeros & zeros \\
        % apply b-dec to input & False & False \\
        % normalize sae decoder & False & False \\
        % scale sparsity penalty by decoder norm & True & True \\
        % decoder heuristic init & True & True \\
        % init encoder as decoder transpose & True & True \\
        % normalize activations & expected average only in & expected average only in \\
        context size & 4096 & 2048 \\
        \bottomrule
    \end{tabular}
    \caption{\textbf{Hyperparameters of training \old{} and \ours{} models.}}
    \label{tab:saetrain}
\end{table}

The hyperparameters of the training \old{} and \ours{} are shown in Table~\ref{tab:saetrain}. The differences in training parameters arise because the LLaVA-NeXT-7B model requires more GPU memory to handle vision input, so fewer batches can be cached. For the \old{} and \ours{} parameters, we set different hook layers and context sizes based on the distinct architectures of the two models. We also experimented with different feature numbers on both models, but found that only around 30,000 features are actually activated during training. All training runs were conducted until convergence. All \old{} and \ours{} training is performed on 8×A800 GPUs and each training typically takes around $21$ hours. We ensured that the variations in the parameters did not affect the experiment results.

% \subsection{Complete Result of \ours{} Capability}
% \label{appendix:reca}
% \begin{table}[h!]
% \centering
% \renewcommand{\arraystretch}{1.2} % 调整行距为1.2倍
% \begin{tabular}{|c|c|c|c|}
% \hline
% \textbf{Model} & \textbf{Method}  & \textbf{$\mathcal{L}_0$ } & \textbf{Reconst. } \\ \cline{3-4}
% \hline
% \multirow{4}{*}{\shortstack{\textbf{LLaVA-}\\\textbf{NeXT-}\\\textbf{7B}}} & Zero &  -  &  10.37 \\ \cline{2-4}
% & \old{} &   94.5 & 6.90 \\ \cline{2-4}
% & \ours{} &   192.5 &  \textbf{2.64} \\ \cline{2-4}
% & Original &  -  &  2.50 \\ \cline{1-4}
% \multirow{4}{*}{\shortstack{\textbf{Chame}\\\textbf{leon-}\\\textbf{7B}}} & Zero & -  &10.95 \\ \cline{2-4}
% & \old{} & 24757.6   & - \\ \cline{2-4}
% & \ours{} &  50.1   & \textbf{7.828} \\ \cline{2-4}
% & Original & -   & 1.682 \\ \cline{1-4}
% \multirow{4}{*}{\shortstack{\textbf{Anole-}\\\textbf{7B}}} & Zero &  -  &  11.09 \\ \cline{2-4}
% & \old{} & 62.1  &  14.25 \\ \cline{2-4}
% & \ours{} & 50.1  &  \textbf{8.61} \\ \cline{2-4}
% & Original & -  &  7.75 \\ \cline{1-4}
% \end{tabular}
% \caption{Comparison of \old{}\ vs. \ours{} across different models. $\mathcal{L}_0$ indicates the sparsity cost (average activated feature number), and \textbf{Reconst.} shows the reconstruction loss. Each section compares the metrics of Zero (set all activations as zero), \old{}, \ours{}, and the Original reference. }
% \label{tab:sae2saev}
% \end{table}

\section{Details of alignment experiment}
\label{appendix:align}

We present details of alignment experiment in this section, including  algorithms and hyperparameters of algorithms and model training. 

\subsection{Algorithms in alignment experiment}

The complete algorithm of $\mathcal{L}_0$-based Ranking and co-occurring $\mathcal{L}_0$-based Ranking are shown in Algorithm~\ref{alg:l0} and Algorithm~\ref{alg:Coocur}. These two algorithms serve as ablation variants of the cosine similarity score Ranking (shown in Algorithm~\ref{alg:cosim}). The $\mathcal{L}_0$-based Ranking represents a straightforward algorithm that selects data by directly computing the sum of $\mathcal{L}_0$ for each data point. The co-occurring $\mathcal{L}_0$-based Ranking takes an initial step toward cross-modal consideration by only counting features that are activated across both modalities. Building upon these algorithms, we further developed the cosine similarity score Ranking approach.

%这两个算法是cosine ranking的ablation 版本，$\mathcal{L}_0$-based Ranking是一个straightforward的算法，通过直接计算一个data的$\mathcal{L}_0$之和来筛选数据，co-occurring $\mathcal{L}_0$-based Ranking则初步考虑了跨模态的角度，只统计在两个模态东被激活的feature。我们在这两个算法的基础上进一步发展得到了cosine similarity

\begin{algorithm}[ht]
\caption{$\mathcal{L}_0$-based Ranking}
\label{alg:l0}
\begin{algorithmic}
\STATE {\bfseries Require:}  multimodal dataset $\mathcal{D}=\{\bm{d}_i\}_{i=1}^{p} $; \model\ $\mathcal{M_\theta}$; \ours{} $\mathcal{S_\theta}$; features of \ours{}; activation bound\text{:} $\delta$;\ $\mathcal{F_\theta}\text{:}\ \{\bm{f}_k\}_{k=1}^{n}$;
\STATE {\bfseries Ensure:} Ranked data $\mathcal{D}_R$
\FOR {each $\bm{d}_i \in \mathcal{D}$}
    \STATE $Z_i \gets \mathcal{S_\theta}(\mathcal{M_\theta}(\bm{d}_i))$ \;
    \STATE $F_i \gets \{\bm{f}_k \text{: } \exists\ z_j \in Z_i ,\ \mathrm{ s.t.}\ z_j\  \text{activates } \bm{f}_k\}$
    \STATE $\mathcal{L}_{0,i} \gets |F_i| $
\ENDFOR
\STATE $\mathcal{D}_R \gets \textit{Sort}(\mathcal{D}, \{\mathcal{L}_{0,i}\}_{i=1}^n)\ $
\end{algorithmic}
\end{algorithm}

\begin{algorithm}[ht]
\caption{Co-ocurring $\mathcal{L}_0$-based Ranking}
\label{alg:Coocur}
\begin{algorithmic}
\STATE {\bfseries Require:} Text token vocabulary: $\mathcal{T}$; vision token vocabulary: $\mathcal{V}$; multimodal dataset $\mathcal{D}=\{\bm{d}_i\}_{i=1}^{p} $; \model\ $\mathcal{M_\theta}$; \ours{} $\mathcal{S_\theta}$; features of \ours{} $\mathcal{F_\theta}\text{:}\ \{\bm{f}_k\}_{k=1}^{n}$; activation bound\text{:} $\delta$;
\STATE {\bfseries Ensure:} Ranked data $\mathcal{D}_R$ 
\STATE Initialize  coocurrence feature set of data $F_i \gets \varnothing$ \;
\FOR {each $\bm{d}_i \in \mathcal{D}$}
\STATE Initialize activated token set of features $\mathcal{A}_k \gets \varnothing $
    \STATE $H_i \gets \mathcal{M_\theta}(\bm{d}_i)$
    \STATE $Z_i \gets \mathcal{S_\theta}(H_i)$ \;
    \FOR {each $\bm{f}_k \in \mathcal{F}_\theta$}
        \STATE $\mathcal{A}_k \gets \mathcal{A}_k \cup \{ \bm{h}_j \text{:}\ \bm{h}_j \in H_i,\bm{z}_j = \bm{e}_j Z_i, \bm{z}_{jk}>\delta\}$
        \IF{$\mathcal{A}_k \cap \mathcal{T} \neq \varnothing \land \mathcal{A}_k \cap \mathcal{V} \neq \varnothing$}
            \STATE $F_i \gets F_i \cup \{\bm{f}_k\}$
        \ENDIF
    \ENDFOR
\ENDFOR
\STATE $\mathcal{D}_R \gets \textit{Sort}(\mathcal{D}, \{|F_i|\}_{i=1}^n )\ $

\end{algorithmic}
\end{algorithm}

\paragraph{Hyperparameters of Algorithms~\ref{alg:cosim},\ref{alg:l0},\ref{alg:Coocur}}
%把所有算法放出来
The hyperparameters of Algorithms~\ref{alg:cosim},\ref{alg:l0},\ref{alg:Coocur}  are
shown in Table~\ref{tab:algo}. We ensure that all parameters are the same to ensure a fair comparison between the algorithms.
%我们确保了所有参数相同以保证算法之间比较的合理性
\begin{table}[ht]
    \centering
    \scriptsize
    \begin{tabular}{cccc}
        \toprule
        \textbf{Hyper-parameters}   &  \textbf{Cosine similarity} &  \textbf{Coocurrence}& $\mathcal{L}_0$ \\
        \midrule
        top-K & 5 & 5& 5 \\
        text token vocabulary size & 32000 & 32000 & 32000 \\
        vision token vocabulary size & 64 & 64 & 64\\
        activation bound & 1 & 1 & 1 \\
        sample data size & 1000 & 1000 & 1000\\
        \bottomrule
    \end{tabular}
    \caption{\textbf{Hyper-parameters of Algorithm~\ref{alg:cosim},\ref{alg:l0},\ref{alg:Coocur}.}}
    \label{tab:algo}
\end{table}

\subsection{Hyperparameter of Model Training}
\label{appendix:modeltrain}

In this section, we list out the hyperparameters used for model training through SFT and DPO (shown in Table~\ref{tab:modeltrain}). All \old{} training is performed on 8×A800 GPUs. To ensure fair comparison between algorithms, we maintained consistent parameter settings across all experiments. 

%在这个部分，我们列出了我们通过SFT和DPO训练模型的具体参数，我们控制了所有的参数保持相同以保证算法之间比较的公平性

\begin{table}[ht]
    \centering
    \scriptsize
    \begin{tabular}{lcc}
        \toprule
        \textbf{Hyper-parameters} & \textbf{SFT}  & \textbf{DPO} \\
        \midrule
        max length & 4096 & 4096 \\
        per device train batch size & 8 & 8\\
        per device eval batch size & 8 & 8\\
        gradient accumulation steps & 4 & 4\\
        LR scheduler type & cosine & cosine\\
        LR & 1e-6 & 1e-6\\
        warmup steps & 10 & 10\\
        eval steps & 50 & 50\\
        epochs & 3 & 3\\
        val size & 0.1 & 0.1\\
        bf16 & True & True\\
        \bottomrule
    \end{tabular}
    \caption{\textbf{Hyperparameters of SFT training and DPO training.} }
    \label{tab:modeltrain}
\end{table}

\section{Details of Applying \ours{} on Multimodal Data}

In this section, we present implementation details of the \ours{} application experiments. We enumerate 4 image patch selection algorithms employed in this study and provide additional case analyses. These comprehensive results further demonstrate the robust inference capabilities of \ours{}.
%在这一部分我们介绍了\ours{} application 实验中的一些细节，我们列出了该实验中的四个图片patch筛选算法，并且展示给更多的图片筛选的例子，这些内容更坚实地展示了\ours{}在推理端的强大能力

\subsection{Algorithm}
\label{appendix:algo}
The complete algorithms of $\mathcal{L}_0$, $\mathcal{L}_1$, co-occurring $\mathcal{L}_0$, and cosine similarity score methods are shown in Algorithm~\ref{alg:l0patch}, Algorithm~\ref{alg:l1patch}, Algorithm~\ref{alg:occpatch} and Algorithm~\ref{alg:cosipatch}.

%这些算法均采用图片作为输入，输出一个被mask的图片，mask的比例取决于mask rate $\gamma$,这些算法均利用了\ours{}的features的激活情况进行了patch filter，主要的区别在于对feature激活情况的计算方法($\mathcal{L}_0$, $\mathcal{L}_1$)和对跨模态相似性的衡量(co-occurring $\mathcal{L}_0$,  cosine similarity score)
\begin{algorithm}[H]
\caption{$\mathcal{L}_0$ patch filter}
\label{alg:l0patch}
\begin{algorithmic}
\STATE {\bfseries Require:}  Vision token vocabulary: $\mathcal{V}$; image $V$; fixed Prompt $T$; \model\ $\mathcal{M_\theta}$; \ours{} $\mathcal{S_\theta}$; features of \ours{} $\mathcal{F_\theta}\text{:}\{\bm{f}_k\}_{k=1}^{n}$; activation bound $\delta$; mask rate $\gamma$;\ \
\STATE {\bfseries Ensure:} Filtered image $V^\prime$
\STATE Initialize score of each patch $p_i \gets 0$
\STATE $H \gets \mathcal{M}_\theta (T,V)$
\STATE $Z \gets \mathcal{S}_\theta (H)$ 

\FOR {each $\bm{h}_i \in H$}
    \IF{$\bm{h}_i \in \mathcal{V}$}
        \STATE $p_i=\sum_j \mathbf{1}(z_{ij}>\delta)$
    \ENDIF
\ENDFOR
\STATE $K \gets \lfloor \gamma |I| \rfloor$
\STATE $V^\prime \gets \textit{TopK}(v_i \in V) \text{ sorted by } p_i$
\end{algorithmic}
\end{algorithm}

\begin{algorithm}[H]
\caption{$\mathcal{L}_1$ patch filter}
\label{alg:l1patch}
\begin{algorithmic}
\STATE {\bfseries Require:}  Vision token vocabulary: $\mathcal{V}$; image $V$; fixed Prompt $T$; \model\ $\mathcal{M_\theta}$; \ours{} $\mathcal{S_\theta}$; features of \ours{} $\mathcal{F_\theta}\text{:}\{\bm{f}_k\}_{k=1}^{n}$; activation bound $\delta$; mask rate $\gamma$;\ \
\STATE {\bfseries Ensure:} Filtered image $V^\prime$
\STATE Initialize score of each patch $p_i \gets 0$
\STATE $H \gets \mathcal{M}_\theta (T,V)$
\STATE $Z \gets \mathcal{S}_\theta (H)$ 

\FOR {each $\bm{h}_i \in H$}
    \IF{$\bm{h}_i \in \mathcal{V}$}
        \STATE $p_i=\sum_j (z_{ij})$
    \ENDIF
\ENDFOR
\STATE $K \gets \lfloor \gamma |I| \rfloor$
\STATE $V^\prime \gets \textit{TopK}(v_i \in V) \text{ sorted by } p_i$
\end{algorithmic}
\end{algorithm}

These algorithms take images as input and produce masked images, where the masking proportion is determined by the mask rate $\gamma$. All algorithms utilize the activation patterns of \ours{} features for patch filtering, with their primary distinctions lying in their methods of computing feature activation ($\mathcal{L}_0$, $\mathcal{L}_1$) and measuring cross-modal similarity (co-occurring $\mathcal{L}_0$, cosine similarity score).

\begin{algorithm}[H]
\caption{$\text{Co-occuring}\ \mathcal{L}_0$ patch filter}
\label{alg:occpatch}
\begin{algorithmic}
\STATE {\bfseries Require:} Text token vocabulary $\mathcal{T}$; vision token vocabulary: $\mathcal{V}$; image $V$; fixed Prompt $T$; \model\ $\mathcal{M_\theta}$; \ours{} $\mathcal{S_\theta}$; features of \ours{} $\mathcal{F_\theta}\text{:}\{\bm{f}_j\}_{j=1}^{n}$; activation bound $\delta$; mask rate $\gamma$;\ \
\STATE {\bfseries Ensure:} Filtered image $V^\prime$
\STATE Initialize score of each patch $p_i \gets 0$, co-occuring feature set $F \gets \varnothing\ \text{and activated token set of features } \mathcal{A}_j \gets \varnothing$
\STATE $H \gets \mathcal{M}_\theta (T,V)$
\STATE $Z \gets \mathcal{S}_\theta (H)$ 
\FOR {each $\bm{f}_j \in \mathcal{F}_\theta$}
    \STATE $\mathcal{A}_j \gets \mathcal{A}_j \cup \{ \bm{h}_i \text{:}\ \bm{h}_i \in H,\bm{z}_i = \bm{e}_i Z, \bm{z}_{ij}>\delta\}$
    \IF{$\mathcal{A}_j \cap \mathcal{T} \neq \varnothing \land \mathcal{A}_j \cap \mathcal{V} \neq \varnothing$}
        \STATE $F \gets F \cup \{\bm{f}_j\}$
    \ENDIF
\ENDFOR
\FOR {each $\bm{h}_i \in H$}
    \IF{$\bm{h}_i \in \mathcal{V}$}
        \STATE $p_i=\sum_j \mathbf{1}(z_{ij}>\delta\ \land\ \bm{f}_j \in F  )$
    \ENDIF
\ENDFOR
\STATE $K \gets \lfloor \gamma |I| \rfloor$
\STATE $V^\prime \gets \textit{TopK}(v_i \in V) \text{ sorted by } p_i$
\end{algorithmic}
\end{algorithm}

\begin{algorithm}[H]
\caption{$\text{Cosine similarity score}$ patch filter}
\label{alg:cosipatch}
\begin{algorithmic}
\STATE {\bfseries Require:} Text token vocabulary $\mathcal{T}$; vision token vocabulary: $\mathcal{V}$; image $V$; fixed Prompt $T$; \model\ $\mathcal{M_\theta}$; \ours{} $\mathcal{S_\theta}$; features of \ours{} $\mathcal{F_\theta}\text{:}\{\bm{f}_j\}_{j=1}^{n}$; activation bound $\delta$; mask rate $\gamma$;cosine similarity weight $\{\omega_j\}_{j=1}^n$\ \
\STATE {\bfseries Ensure:} Filtered image $V^\prime$; 
\STATE Initialize score of each patch $p_i \gets 0$, co-occuring feature set $F \gets \varnothing\ \text{and activated token set of features } \mathcal{A}_j \gets \varnothing$
\STATE $H \gets \mathcal{M}_\theta (T,V)$
\STATE $Z \gets \mathcal{S}_\theta (H)$ 
\FOR {each $\bm{f}_j \in \mathcal{F}_\theta$}
    \STATE $\mathcal{A}_j \gets \mathcal{A}_j \cup \{ \bm{h}_i \text{:}\ \bm{h}_i \in H,\bm{z}_i = \bm{e}_i Z, \bm{z}_{ij}>\delta\}$
    \IF{$\mathcal{A}_j \cap \mathcal{T} \neq \varnothing \land \mathcal{A}_j \cap \mathcal{V} \neq \varnothing$}
        \STATE $F \gets F \cup \{\bm{f}_j\}$
    \ENDIF
\ENDFOR
\FOR {each $\bm{h}_i \in H$}
    \IF{$\bm{h}_i \in \mathcal{V}$}
        \STATE $p_i=\sum_j \mathbf{1}(z_{ij}>\delta\ \land\ \bm{f}_j \in F  )\ \omega_j$
    \ENDIF
\ENDFOR
\STATE $K \gets \lfloor \gamma |I| \rfloor$
\STATE $V^\prime \gets \textit{TopK}(v_i \in V) \text{ sorted by } p_i$
\end{algorithmic}
\end{algorithm}

\subsection{Case Analysis}
\label{appendix:Semantic_Experiment}

\begin{figure}[H]
    \centering
    \includegraphics[width=1\linewidth]{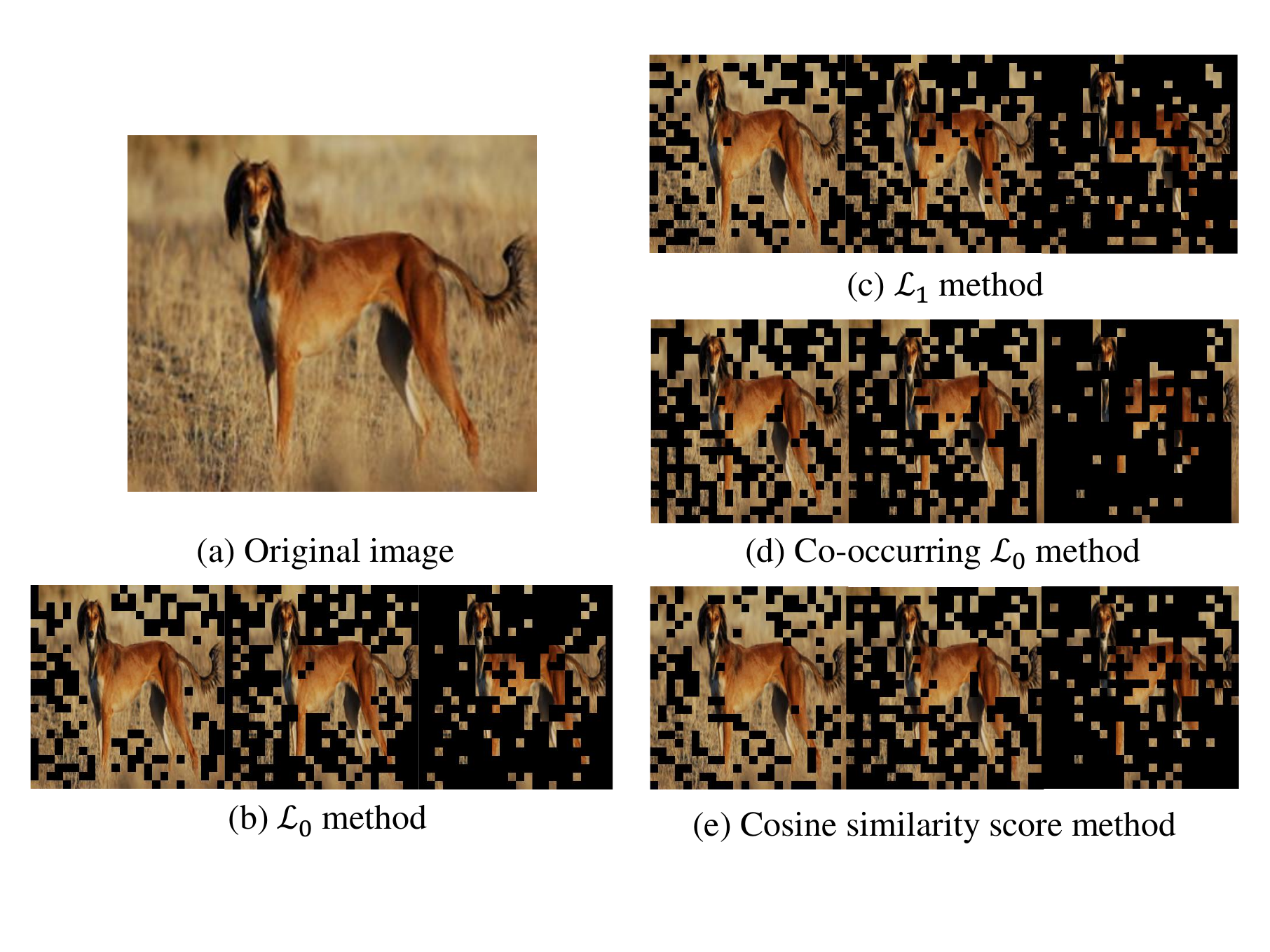}
    \vspace{-0.6cm}
    \caption{\textbf{Case Analysis of all image patch filtering methods in Section~\ref{sec:Semantic}.} We present the original image (a) and 4 case for methods, $\mathcal{L}_0$ (b), $\mathcal{L}_1$ (c), co-occurring $\mathcal{L}_0$ (d) and cosine similarity score (e). Each case contains 3 images as preserving top 75\% patches, top 50\% patches and top 25\% patches. }
    \label{fig:allmethod}
\end{figure}

We present 4 cases in Figure~\ref{fig:allmethod}, corresponding to each of our metric in Section~\ref{sec:Semantic}. The cases intuitively show that $\mathcal{L}_0$ method and cosine similarity score method are more capable of identifying significant patches in images compared to other methods, which aligns with the quantized results shown in Figure~\ref{fig:seminar_experi_2}.

\begin{figure}[H]
    \centering
    \includegraphics[width=1\linewidth]{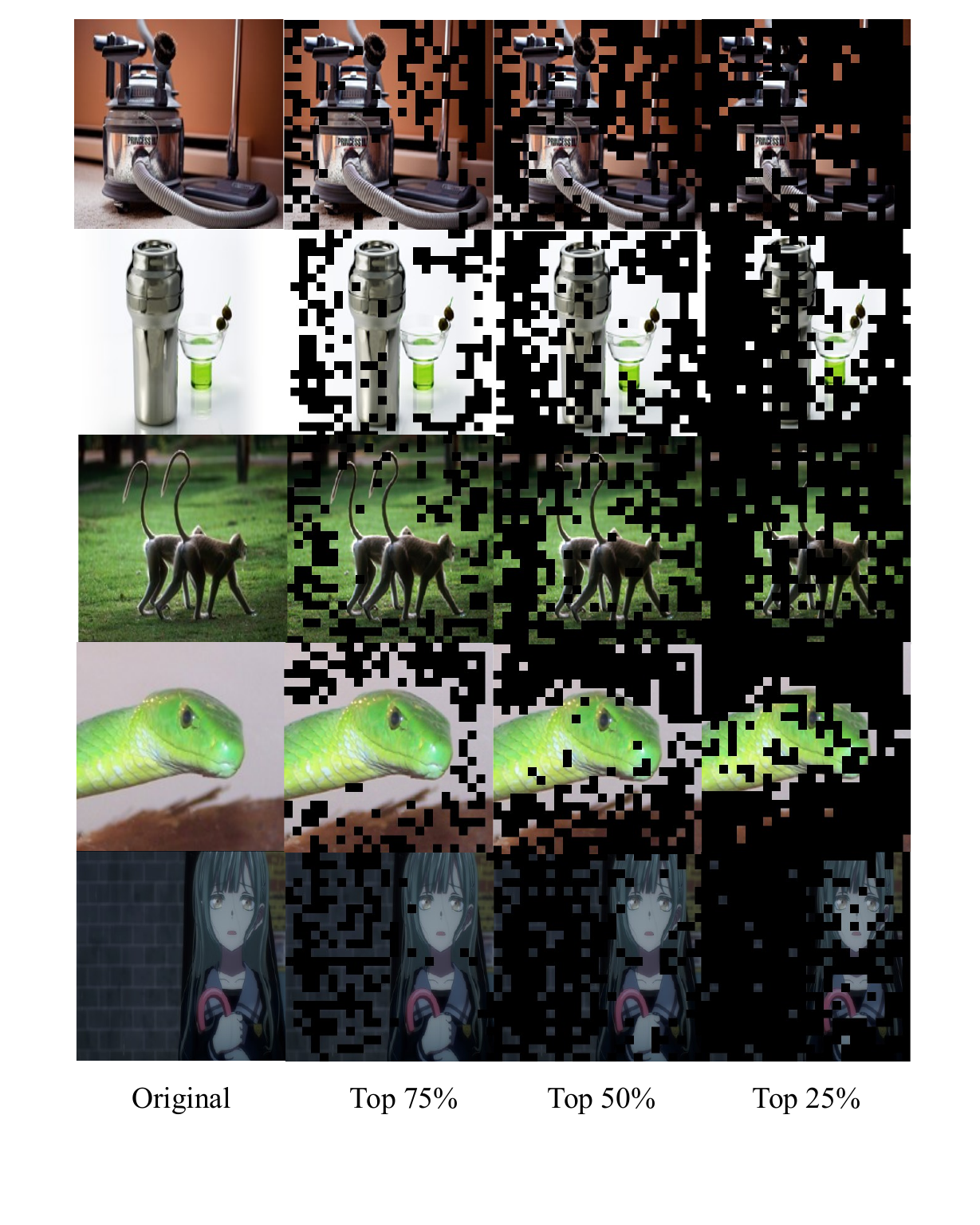}
    \vspace{-1cm}
    \caption{\textbf{Case Analysis of cosine similarity score method in Section~\ref{sec:Semantic}.} 5 cases filtered with cosine similarity score method are shown in the Figure. Each case contains contains 4 images as original image, preserving top 75\% patches, top 50\% patches and top 25\% patches.}
    \label{fig:cosi}
\end{figure}

We present 5 cases filtered with the cosine similarity score method in Figure~\ref{fig:cosi}. The results show that \ours{} model performs excellently in capturing critical patches in images.

\subsection{Quantative Analysis}
\label{appendix:quant_anal}

\begin{table}[htbp]
\centering
\begin{tabular*}{\columnwidth}{@{\extracolsep{\fill}}lcccc}
\toprule
\textbf{Masking (\%)} & \textbf{0} & \textbf{25} & \textbf{50} & \textbf{75} \\
\midrule
\textbf{Text Accuracy (\%)} & 80.2 & 77.7 & 66.5 & 53.3 \\
\textbf{Image Accuracy (\%)} & 80.2 & 78.8 & 78.4 & 70.7 \\
\bottomrule
\end{tabular*}
\label{tab:masking_accuracy}
\caption{\textbf{Applying \ours{} on VQA tasks} We apply \ours{}-based patch filtering experiment on the text and image of VQA task separately, using A-OKVQA validation set with LLaVA-NeXT-7B model. The experiment shows that image information demonstrates a lower compression rate (more redundancy) than text. Moreover, text masking shows a roughly linear relation of accuracy and masked token, while image filter maintains performance until 50\%, with a more significant drop only appearing at 75\%, suggesting that the information of text is more evenly distributed compared to image.}
\end{table}

We performed additional experiments on \ours{}-based patch filtering. To be specific, we use \ours{}-based patch filter on the text and image part of each VQA question separately, and test the accuracy when part of the information is masked. The results show that image information demonstrates a lower compression rate (more redundancy) than text. Moreover, text masking shows a roughly linear relation of accuracy and masked token, while image filter maintains performance until 50\%, with a more significant drop only appearing at 75\%, suggesting that the information of text is more evenly distributed compared to image.

\end{document}